\documentclass[10pt,twocolumn,letterpaper]{article}

\usepackage[pagenumbers]{cvpr} 

\newcommand{\supparxiv}[2]{#2}

\supparxiv{

}{

}

\newcommand{\newpara}[1]{\vspace{6pt}\noindent\textbf{#1}}

\usepackage{pifont}

\newcommand{\xmark}{\ding{55}}  

\usepackage{titletoc}
\usepackage{tocloft}
\usepackage[table,xcdraw]{xcolor}
\usepackage{makecell,array}

\definecolor{lightergray}{gray}{0.93}

\definecolor{cvprblue}{rgb}{0.21,0.49,0.74}
\usepackage[pagebackref,breaklinks,colorlinks,allcolors=cvprblue]{hyperref}
\usepackage[accsupp]{axessibility}  

\input{def_custom.tex}
\def\paperID{43611} 
\def\confName{CVPR}
\def\confYear{2026}

\title{Seeing Through Touch: Tactile-Driven Visual Localization of Material Regions}

\author{Seongyu Kim$^{1}$ \quad Seungwoo Lee$^{1}$ \quad Hyeonggon Ryu$^{2}$ \quad Joon Son Chung$^{1}$ \quad Arda Senocak$^{3}$\\
\\
$^{1}$Korea Advanced Institute of Science and Technology \hspace{5pt}
$^{2}$Hankuk University of Foreign Studies
\\
$^{3}$Ulsan National Institute of Science and Technology
}

\begin{document}
\twocolumn[{%
\renewcommand\twocolumn[1][]{#1}%
\maketitle
\begin{center}
\centering
\vspace{-0.5em}
\captionsetup{type=figure}
\includegraphics[width=\linewidth]{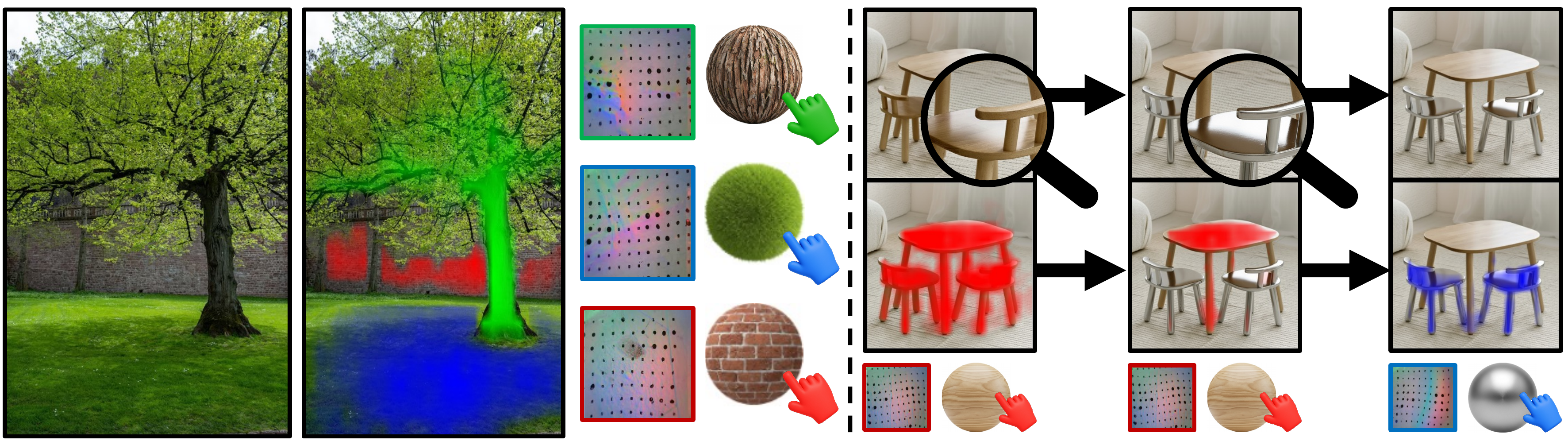}
\caption{\textbf{Tactile Localization.} We introduce a task where the goal is to localize the visual regions that correspond to given tactile inputs. 
Left: The scene remains the same while different touch signals are given, and the model localizes the corresponding regions. Right: Some regions are replaced with different materials; the model should stop highlighting them when they no longer match the touch signal and highlight them again once the touch signal matches the new material.
}
\label{fig:teaser}
\end{center}
}]
\begin{abstract}
We address the problem of tactile localization, where the goal is to identify image regions that share the same material properties as a tactile input. Existing visuo-tactile methods rely on global alignment and thus fail to capture the fine-grained local correspondences required for this task. The challenge is amplified by existing datasets, which predominantly contain close-up, low-diversity images. We propose a model that learns local visuo-tactile alignment via dense cross-modal feature interactions, producing tactile saliency maps for touch-conditioned material segmentation. To overcome dataset constraints, we introduce: (i) in-the-wild multi-material scene images that expand visual diversity, and (ii) a material-diversity pairing strategy that aligns each tactile sample with visually varied yet tactilely consistent images, improving contextual localization and robustness to weak signals. We also construct two new tactile-grounded material segmentation datasets for quantitative evaluation. Experiments on both new and existing benchmarks show that our approach substantially outperforms prior visuo-tactile methods in tactile localization. Project
page: \url{https://mm.kaist.ac.kr/projects/SeeingThroughTouch/}.
\end{abstract}    
\vspace{-6mm}
\section{Introduction}
\label{sec:intro}
Humans possess a natural ability to infer the tactile properties of the world around them~\cite{graziano1995representation}. With a single touch, we can grasp how a material feels, its softness, roughness, or texture, and immediately associate that sensation with visual cues in the environment~\cite{smith2005development}. When we touch a velvet cloth, we can easily identify other regions in a scene that would evoke the same tactile feeling, even without physically touching them. This cross-modal skill suggests that tactile perception and visual understanding are intertwined, allowing us to reason about how things feel simply by looking at them.

Inspired by humans’ ability to imagine how surfaces feel from their visual appearance, we investigate whether machines can perform a similar reasoning process. To this end, we define the \textit{tactile localization} task, where a model is given a tactile input and must identify regions in an image that share similar material properties. This task can be viewed as a material segmentation conditioned on tactile cues rather than purely visual ones~\cite{bell2015material,upchurch2022dense,sharma2023materialistic}, guiding the model to learn visual features that correspond to tactile properties. However, enabling such visuo-tactile reasoning poses several unique challenges.

We focus on learning local visuo-tactile representations that emerge from dense cross-modal feature interactions, a capability often missing in existing approaches. Prior work on visuo-tactile learning has predominantly focused on global alignment between modalities, using similarity between pooled representations or CLS tokens to capture coarse semantic correspondence across entire samples. While such methods can determine whether an image and a tactile input correspond to the same material, they fail to identify where in the visual scene a given tactile property exists. This limitation prevents existing models from supporting downstream tasks that require spatial reasoning, such as tactile localization. To address this, we propose a model that computes dense similarity maps between local tactile and visual features, producing tactile saliency maps that highlight image regions expected to evoke the same tactile sensation as the given touch.

In addition to architectural limitations, existing dataset constraints also make this task challenging to learn. Most existing datasets consist of close-up, texture-centric images in which nearly the entire image corresponds to a single tactile category, as the contact point or object is shown at a very close range. Additionally, the visual frames corresponding to the touch signals remain nearly identical (see~\Fref{fig:limited_image}). This limited information and diversity result in only a few effective image–tactile pairs, making them insufficient for learning strong cross-modal alignment. To address these challenges, we adopt two key strategies. First, we extend the visual data with in-the-wild, open-world, scene-level images containing multiple material types. Second, leveraging the observation that similar materials evoke similar tactile sensations, we introduce a material diversity-based pairing strategy that associates one tactile sample with multiple visually diverse yet tactilely consistent images. This not only enriches the visuo-tactile correspondence space, allowing the model to learn more robust contextual tactile localization, but also provides an emergent ability to handle weak tactile signals more effectively.

Lastly, the main goal of this paper is to achieve tactile-grounded material segmentation; accordingly, we aim to evaluate our model’s ability to perform this task. However, no existing dataset provides both image-tactile pairs and corresponding segmentation maps suitable for this purpose. Therefore, we constructed two new datasets: the first extends the Touch-and-Go (TG)~\cite{yang2022touch} dataset by adding pixel-level material segmentations for the corresponding touch samples, and the second is a newly curated dataset of scene-level in-the-wild images collected from the web, annotated with material regions aligned to tactile categories. Evaluation on both new and existing datasets, including OpenSurfaces~\cite{bell2013opensurfaces}, shows that our model outperforms prior visuo-tactile methods and baselines in tactile localization.

Our main contributions are as follows:
\begin{itemize}
    \item We propose a local visuo–tactile alignment model that produces dense tactile saliency maps to identify image regions sharing the same tactile sensation as a given touch input for material segmentation.
    \item We curate in-the-wild, scene-level multi-material images and propose a material-diversity pairing strategy that enriches local visuo-tactile correspondence and improves robustness to weak tactile signals.
    \item We construct two new tactile-grounded material segmentation datasets for evaluation and  demonstrate that our model outperforms existing methods.

\end{itemize}
\section{Related Work}\label{sec:related}
\newpara{Visuo-Tactile Representation Learning.} Early work on modeling cross-modal associations between touch and vision jointly learned a shared representation by training CNNs across visual, tactile, and depth images for fabric classification~\cite{yuan2017connecting}. More recently, this direction has shifted toward self-supervised learning~\cite{zambelli2021learning,yang2022touch,kerr2022self,yang2024binding,dou2024tactile,fu2024a,dave2024multimodal,gungor2025towards}. While~\cite{yang2022touch,kerr2022self} employ general multimodal contrastive learning between vision and touch,~\cite{dave2024multimodal} extends this idea by incorporating both inter- and intra-modal relationships. Further studies~\cite{yang2024binding} expand beyond touch–vision learning by connecting the tactile modality to language and sound as well, by aligning touch embeddings with image embeddings~\cite{girdhar2023imagebind} already aligned with language and audio. Likewise,~\cite{fu2024a} connects touch to vision and language but adopts a pairwise approach rather than binding via images. Although these methods address visuo-tactile learning, they primarily emphasize global alignment, producing embeddings that capture coarse semantic correspondences across entire samples but lack localization ability in their local features. In contrast, our work focuses on fine-grained local alignment, where tactile sensations correspond to specific visual regions that feel similar, using contrastive learning at the local level.
\begin{figure*}[t!]
    \centering
    \includegraphics[width=0.98\linewidth]{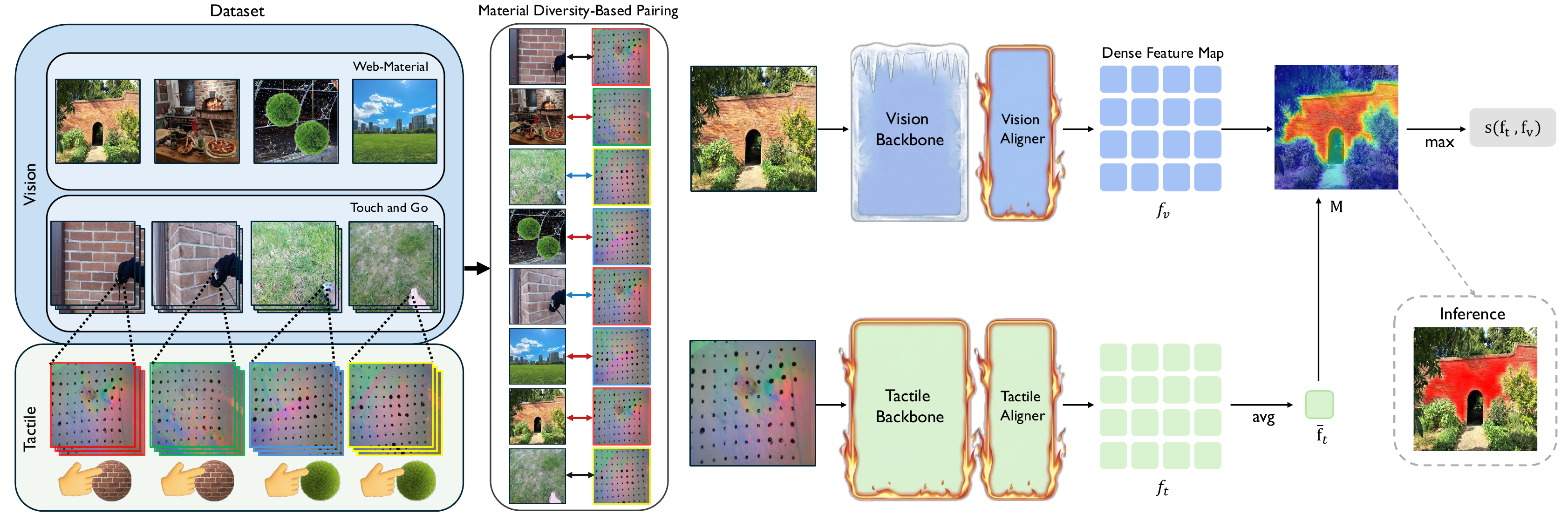}
    \caption{\textbf{Pipeline of \textit{Seeing Through Touch}.} Tactile and visual encoders extract features from touch signals and paired images. These features are used to compute visuo-tactile similarities for contrastive learning. We extend visuo-tactile pairing by linking each tactile signal to diverse \textcolor{in_domain}{in-domain} (Touch-and-Go) and \textcolor{out_domain}{out-domain} (Web-Material) images of the same material category, leveraging the insight that similar materials evoke similar tactile sensations.}
    \label{fig:pipeline}
\end{figure*}

\newpara{Visuo-Tactile Localization.} Tactile localization has been explored in various forms across the field, with differing objectives but a shared formulation in which, given a tactile signal, the goal is to identify the corresponding region in a visual scene. ~\cite{kerr2022self} defines this task as contact localization on garment data using a robotic arm, following pre-training with spatially aligned tactile–vision data. TaRF~\cite{dou2024tactile} integrates tactile sensing into neural radiance fields to learn a shared 3D representation of vision and touch, enabling spatially aligned tactile prediction within 13 reconstructed scenes and localization of contact points in 3D space. In contrast, our work focuses on tactile localization in in-the-wild 2D RGB images, where no explicit geometry or scene reconstruction is available and the domain is not restricted to garments. Instead of pinpointing a single contact location, our model segments all image regions that exhibit similar material properties to the given touch. From this perspective, our task is related to~\cite{sharma2023materialistic}, which selects image regions sharing the same material with a user-selected query pixel. However, unlike~\cite{sharma2023materialistic}, our method uses tactile sensory input as the query rather than a pixel within the same image, making the problem more challenging, as it requires discovering cross-modal correspondences rather than performing in-domain segmentation.

\section{Methodology}\label{sec:method}

Our goal is tactile localization -- identifying image regions that share the same tactile sensation as a given touch input for material segmentation. We propose \textit{Seeing Through Touch (STT)}, a framework that encodes paired tactile and visual inputs into a shared space and learns fine-grained local cross-modal alignment through contrastive learning. To enhance alignment, a material diversity-based pairing strategy leverages intra-category material variation, while additional in-the-wild web images further improve generalization. An overview is shown in Figure~\ref{fig:pipeline}.

\subsection{Preliminaries}

\newpara{Contrastive Learning} encourages positive pairs to be close and negative pairs to be far apart. In visuo-tactile learning, let $E_t$ and $E_v$ be the tactile and visual encoders, respectively. Given a tactile feature ${f}_{t_i}=E_t(t_i)$ and its positive visual counterpart ${f}_{v_i}=E_v(v_i)$, with negatives ${f}_{v_j}$ ($i\neq j$) from a dataset $\calD=\{(v_i,t_i)\}_{i=1}^N$, the loss is:
\begin{equation}
\label{eq:tv_contrastive}
\calL_i = -\log \frac{\exp(s(f_{t_i}, f_{v_i})/\tau)}{\sum_j \exp(s(f_{t_i}, f_{v_j})/\tau)},
\end{equation}

\noindent where $s$ is a cross-modal similarity and $\tau$ is a temperature~\cite{wu2018unsupervised}. As in prior works~\cite{senocak2023sound,radford2021learning,mo2022localizing,yang2022touch,yang2024binding,ryu2025seeing}, this loss is applied symmetrically.

\newpara{Vision and Tactile Encoders with Aligner.} Given an image $v_i$ and its paired tactile sample $t_i$, the encoders extract modality-specific features. Both encoders consist of a backbone network followed by a shallow aligner network. This process maps the visual and tactile inputs into a shared representation space, producing a visual feature map $f_{v} \in \mathbb{R}^{C \times H \times W}$ and a tactile feature map $f_{t} \in \mathbb{R}^{C \times H \times W}$. Here, $H$ and $W$ denote the spatial dimensions, and $C$ represents the channel dimension of the shared feature space.

\newpara{Similarity Function.} To achieve fine-grained visuo-tactile alignment, it is essential to use a function that computes the similarity between tactile and visual features with consideration of the task. As we aim to find the regions in the image that correspond to the given tactile input, we first aggregate the tactile feature into a 1-D vector:
\begin{align}
\bar{f}_{t} &= \text{avg}_{h, w} \left(f_{t}[h, w]\right),\label{eq:t_aggregate}
\end{align}
where $f_{t}[h, w] \in \mathbb{R}^{C}$ refers to the 1-D vector at location $[h, w]$ of $f_{t} \in \mathbb{R}^{C\times H\times W}$.
We then construct a similarity map $ M \in \mathbb{R}^{H \times W}$ from the aggregated tactile feature and the visual feature map:
\vspace{-2mm}
\begin{align}
M[h,w] = \bar{f}_{t} \cdot f_{v}[h, w],\label{eq:similarity_map}
\end{align}
where $f_{v}[h, w] \in \mathbb{R}^{C}$ denotes the 1-D vector at location $[h, w]$, and · represents the inner product. Thus, we obtain a similarity map between the tactile input and its corresponding image. The final similarity score is the max-pooled value of the similarity map:
\begin{align}
s(f_{t},f_{v}) = \max(M).\label{eq:similarity_score}
\end{align}

\subsection{Training Pairs}\label{sec:ori_pair} 
By analyzing the popular visuo-tactile benchmark TG~\cite{yang2022touch}, we make several observations and use them to design our training pair strategy as follows.

\newpara{Touch Instance.} 
A touch instance refers to the action in which the collector presses and releases the sensor on an object surface, during which synchronized tactile and image frames are recorded~\cite{yang2022touch}. Formally, it consists of a sequence of frames $(v_1, t_1), (v_2, t_2), \dots, (v_T, t_T)$, where $T$ is the number of frames in the instance. Although tactile signals vary throughout a touch instance, the corresponding visual frames remain nearly identical despite slight camera pose changes, making the visual modality temporally invariant.

\newpara{Positive Pair Construction.} 
Images and tactile signals from the same touch instance inherently correspond to the same material. Thus, any tactile frame $t_i$ and image frame $v_i$ from that instance can form a positive pair, as done in prior contrastive learning–based methods. However, given our observation that the visual modality is temporally invariant, positive pairs can also be constructed by randomly sampling a tactile frame $t_j$ and an image frame $v_i$ from the same instance, even when $i \neq j$. The similarity between tactile and visual features is then computed using Eq.~\ref{eq:similarity_score} as $s(f_{t_j}, f_{v_i})$, which represents our training pair strategy. Moreover, this temporal invariance of the visual modality enables leveraging the fact that similar materials evoke similar tactile sensations. As a result, tactile frames can be mapped to visually similar images without requiring precise temporal correspondence, as we discuss in the next section.

\subsection{Material Diversity-Based Pairing}\label{sec:multi_positive} 
In the visuo-tactile context, a tactile sensation is treated as a positive pair with its corresponding image, while negative pairs are sampled from other images. However, as mentioned earlier, the images within the same touch instance are highly similar, so regardless of the pairing combinations within each instance, only a few effective image–tactile pairs are obtained. This limited diversity makes the learning objective insufficient for strong cross-modal alignment. To address this, we extend contrastive learning by pairing each tactile signal with diverse images from both in-domain and out-domain samples of the same material category, thereby improving visuo-tactile alignment.

\newpara{In-domain Pairing.} 
Let our dataset consist of $N$ touch instances, each represented as $(c_n, y_n)$, where $c_n$ is the $n$-th instance and $y_n$ is its material category label:  $\mathcal{D} = \{(c_1, y_1), (c_2, y_2), \dots, (c_N, y_N)\}$. 
Each instance $c_n$ contains a sequence of synchronized tactile and image frames: $c_n = \{({v}_1^n, {t}_1^n), ({v}_2^n, {t}_2^n), \dots, ({v}_{T_n}^n, {t}_{T_n}^n)\}$. 
Even within the same material category $y_n$, tactile signals can vary due to inherent structural differences between instances. To capture this diversity, we consider all instances that share the same material category and randomly sample tactile and image frames across these instances to construct positive pairs. The similarity between a tactile feature $f_{t_j}^n$ from instance $c_n$ and an image feature $f_{v_i}^m$ from instance $c_m$ of the same material category is computed using Eq.~\ref{eq:similarity_score} as $s(f_{t_j}^n, f_{v_i}^m)$.

\newpara{Out-domain Pairing.}
Let the additional image dataset be $D_\text{out} = \{(v_1, y_1), (v_2, y_2), \dots, (v_l, y_l)\}$ 
where each image $v_i$ is labeled with its material category $y_i$. Since these images do not have temporal sequences and are not naturally paired with tactile data, we construct positive pairs by sampling tactile frames $t_j^n$ from existing instance $c_n$ in the main dataset whose material category $y_n$ matches $y_i$. The similarity function between a tactile feature and an out-domain image feature is computed using Eq.~\ref{eq:similarity_score} as $s(f_{t_j}^n, f_{v_i})$. 
This formulation allows all $l$ images in $D_\text{out}$ to be leveraged for cross-modal alignment, using tactile frames drawn from the existing dataset.

\subsection{Collecting Additional Images}\label{sec:image_collection}
In the Touch-and-Go dataset, the images are extremely close-up, with the material filling almost the entire scene except for the collector's hand and the sensor, as illustrated in~\Fref{fig:limited_image}. Consequently, the dataset is ineffective as both a training source and an evaluation benchmark for tactile localization. To address this, we collect additional scene-level images containing multiple material types from the web and a prior material understanding dataset~\cite{bell2015material}. As described in~\Sref{sec:multi_positive}, these open-world images are paired with tactile samples from the TG based on material categories, since similar materials evoke similar tactile sensations.

\newpara{Image Collection.}
We collect images from search engines using descriptive phrases that capture diverse real-world contexts for each material. For every tactile category in the TG dataset, we prompt an LLM~\cite{comanici2025gemini,achiam2023gpt} to generate richer queries beyond simple class names, which are then used to retrieve relevant and diverse web images. The LLM instruction for this process is provided in the supplementary material. For example, for the category “Brick”, the LLM generates phrases such as “brick house in a suburban neighborhood”, “brick chimney in a cozy living room”, and “brick bridge over a river”. By incorporating such phrases covering varied objects, structures, and environments, the collected images represent each material across a broad range of visual and contextual variations. Additionally, we collect images from MINC~\cite{bell2015material}, a dataset of materials in the wild. Since we use tactile data from the TG dataset, 11 of its 18 categories overlap with MINC, yielding about 17K samples.

\newpara{Image Filtering.}  
After collecting images from the web, we filter out misclassified samples belonging to visually similar but incorrect categories. For each category, the LLM suggests potentially confusing alternatives, and we compute a CLIP~\cite{radford2021learning} similarity score between images and category names, retaining only images whose original category has the highest similarity score. For example, for “Brick”, we use prompts such as “a photo of brick”, “a photo of concrete”, “a photo of stone”, and “a photo of rock”, keeping only images most similar to the first prompt. After this automated filtering, human annotators make minimal refinements by removing remaining unrelated images.

\subsection{Implementation and Training Details}

\subsubsection{Architecture Details}

\newpara{Image encoder $E_v(\cdot)$ and Tactile encoder $E_t(\cdot)$.} We adopt DINOv3~\cite{simeoni2025dinov3}, pre-trained on large scale image datasets~\cite{deng2009imagenet, russakovsky2015imagenet,warburg2020mapillary}, in a self-supervised manner, for both encoders. An aligner module consisting of a Channel-wise LayerNorm~\cite{ba2016layernormalization} block and a 1x1 convolutional layer is appended to the end of each backbone. During training, we freeze  the image backbone and optimize the image aligner together with the entire tactile encoder according to our training objective.

\subsubsection{Implementation Details}
Our model takes a single image and a tactile signal as input, each of size 224 × 224. Both image and tactile data are preprocessed with standard image augmentations. The training is conducted on 4 A5000 GPUs with an effective batch size of 64. See the supplementary material for details.
\section{Experiments}
\label{sec:experiments}
\subsection{Datasets}\label{sec:datasets}
\newpara{Training datasets.}  
Our training data come from two sources: 1) \textbf{Touch-and-Go (TG)}~\cite{yang2022touch}: It contains  approximately 246k pairs of images and tactile signals. Although several visuo–tactile datasets exist~\cite{kerr2022self,gao2022objectfolder,gao2023objectfolder,li2019connecting, calandra2017feeling}, most are limited to tabletop, object-centric, 
or simulated settings. Since we target tactile localization in real scenes, we use TG, which was collected by human operators across diverse sub-scene environments beyond controlled setups. 2) \textbf{Our image dataset (\Sref{sec:image_collection}}): It consists of 32,107 web-collected images with diverse scenes and multiple material categories. 
Based on the training setup described in~\Sref{sec:ori_pair} and~\Sref{sec:multi_positive}, these datasets are used accordingly.

\newpara{Testing datasets.} We evaluate localization performance using the following datasets. As no existing dataset fits this purpose, we created new benchmarks.

\begin{itemize}
\item \textbf{TG-Test:} We manually annotated material segmentation masks based on ground-truth tactile categories. Since this dataset already contains tactile signals, it naturally forms visuo-tactile pairs. The pairs are taken from the test split of~\cite{yang2022touch}, totaling 579 samples across 18 categories.
\item \textbf{Web-Material:} We manually annotated material segmentation masks for web-crawled images that are completely disjoint from the training split. Tactile signals are mapped from the Touch-and-Go dataset based on category matching, yielding 675 samples covering 18 categories.
\item \textbf{OpenSurfaces}~\cite{bell2013opensurfaces}\textbf{:} As it already provides segmentation masks for material recognition, we use them directly and map tactile signals from TG based on overlapping categories, resulting in 211 samples across 13 categories. As these datasets lack corresponding tactile signals, we pair each image with a prototype feature computed by averaging the start, middle, and end tactile frames.
\end{itemize}

\noindent We use an online annotation tool~\cite{cvat} built on top of the Segment Anything Model (SAM)~\cite{kirillov2023segment}. By selecting keypoints on images with simple mouse clicks, annotators obtain high-quality segmentation masks. They segment each region according to the given tactile categories. Example annotations are shown in~\Fref{fig:qualitative_results}.

\subsection{Baselines}\label{sec:baselines}
We compare our methods against the baselines and prior visuo-tactile works, grouping them according to the perspectives used for analysis and comparison.

\newpara{Visual Bias.}
As mentioned earlier, the visual images may consist of close-up shots or exhibit visual bias, as the region of interest is often single, centered, or easily identifiable without tactile information. We introduce the following baselines to assess and reveal such bias: (1) \textit{Full Square} and (2) \textit{Full Circle} binary masks: These involve no visual or tactile understanding and simply apply a fixed square (224×224) or circular (diameter 224) mask; (3) \textit{DINOv3 Attention Map}: a vision-only baseline that captures visual objectness without tactile cues.

\newpara{Global vs. Local Alignment.}
Existing visuo-tactile works primarily employ global alignment between the two modalities, whereas localization requires dense, fine-grained local alignment. The following baselines are used to validate this hypothesis and our learning objective: 
(1) \textit{TVL}~\cite{fu2024a}: A recent state-of-the-art visuo-tactile method that aligns CLS tokens. We use its without-language setting for fair comparison. A variant with a frozen CLIP-Large pretrained image encoder and a ViT-Tiny tactile encoder is trained from scratch using the same data as ours. 
(2) \textit{STT-CLS}: A variant of our model trained with a CLS-token alignment objective. 
(3) \textit{STT}-Local: A variant of our model trained with a proposed local alignment objective only with positive pair construction of \Sref{sec:ori_pair}.
(4) \textit{STT}-Indomain: A version of \textit{STT}-Local that incorporates In-domain material diversity-based pairing.  
(5) \textit{Seeing Through Touch (STT)}: Our final model, which extends \textit{STT}-Local by employing Out-domain material diversity-based pairing.

\newpara{Upper Bound Baselines.}
We also include two upper-bound references. These baselines are not intended for direct comparison, but serve as reference points to indicate how far localization performance could reach under more favorable conditions: (1) \textit{GroundedSAM}~\cite{ren2024grounded}: A large-scale vision–language segmentation model that uses text prompts for segmentation; in our experiments, the tactile category name is used as a prompt. (2) \textit{Materialistic}~\cite{sharma2023materialistic}: A material segmentation model that uses a user-clicked visual prompt; in our setup, a pixel from the ground-truth segmented area is provided as the prompt.

\subsection{Main Results}
\begin{figure*}[t!]
    \centering
    \includegraphics[width=1.0\linewidth]{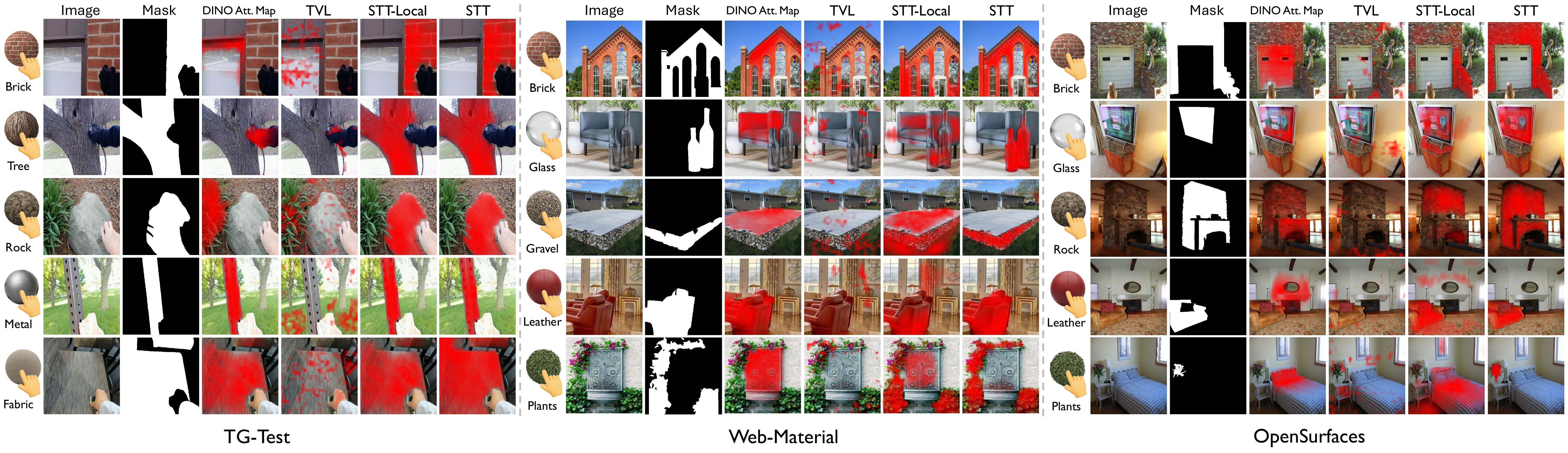}
    \vspace{-7mm}
    \caption{\textbf{Qualitative Tactile Localization Results.} Our model localizes more accurately than prior works and baselines across all benchmarks.}
    \label{fig:qualitative_results}
    \vspace{-4mm}
\end{figure*}

\begin{table}[t]
\centering
\resizebox{\linewidth}{!}{
\begin{tabular}{lcccccc}
\toprule
Model & \multicolumn{2}{c}{TG-Test} & \multicolumn{2}{c}{Web-Material} & \multicolumn{2}{c}{OpenSurfaces~\cite{bell2013opensurfaces}} \\
\cmidrule(lr){2-3} \cmidrule(lr){4-5} \cmidrule(lr){6-7}
& mAP & mIoU & mAP & mIoU & mAP & mIoU \\
\midrule
\rowcolor{lightergray}\textit{Binary-mask} & & & & & & \\
Full Square & - & 67.25 & - & 32.13 & - & 18.13 \\
Full Circle & - & 61.75 & - & 34.20 & - & 18.19 \\
\bottomrule
\rowcolor{lightergray}\textit{Visual Heatmap} & & & & & & \\
DINOv3 Att. Map~\cite{simeoni2025dinov3} & 83.74 & 74.27 & 62.73 & 47.12 & 18.91 & 19.04 \\
\midrule
\rowcolor{lightergray}\textit{Global Alignment} & & & & & & \\
TVL w/o Language~\cite{fu2024a} & 70.61 & 68.12 & 32.16 & 32.16 & 17.93 & 18.61 \\
\textit{STT}-CLS & 73.63 & 73.49 & 39.35 & 34.74 & 17.98 & 19.07 \\
\midrule
\rowcolor{lightergray}\textit{Local Alignment} & & & & & & \\
\textit{STT}-Local & 85.12 & 76.79 & 67.72 & 52.34 & 37.25 & 29.47 \\
\textit{STT}-Indomain & 86.95 & 77.58 & 71.33 & 55.73 & 42.54 & 34.10 \\
\textit{STT} & 87.56 & 76.82 & 77.43 & 60.94 & 48.06 & 36.73 \\
\bottomrule
\rowcolor{lightergray}\multicolumn{3}{l}{\textit{Upper Bound Baselines}} & & & & \\
{GroundedSAM~\cite{ren2024grounded}} & - & 77.22 & - & 67.03 & - & 50.23 \\
{Materialistic~\cite{sharma2023materialistic}}  & 96.29 & 87.91 & 91.22 & 76.14 & 88.77 & 69.83\\
\midrule

\end{tabular}
}
\vspace{-3mm}
\caption{\textbf{Tactile Localization Results on TG-Test, Web-Material, and OpenSurfaces.}}
\label{tab:dataset_all_main}
\vspace{-6mm}
\end{table}

\subsubsection{Comparison with Prior Works and Baselines}
We evaluate our method against prior works and strong baselines on three test sets, TG-Test, Web-Material, and OpenSurfaces, defined in~\Sref{sec:datasets}. We use mAP and mIoU as evaluation metrics following the standard multimodal grounding protocols~\cite{everingham2015pascal,luddecke2022image,cheng2022masked,hamilton2024separating,ryu2025seeing}. The results are presented in~\Tref{tab:dataset_all_main}. Our model consistently outperforms prior works and relevant baselines. \textbf{Key findings} are as follows:

\noindent \textit{1. Local visuo-tactile alignment is essential for tactile localization.} Our results show that global alignment is not a suitable objective for learning tactile localization. All variants of our local alignment objective outperform both TVL and \textit{STT-CLS} across all test sets by a large margin, as these methods rely on global alignment. This validates the motivation of our work and shows the necessity of the proposed method over existing visuo-tactile approaches.

\noindent \textit{2. Material diversity-based pairing effectively improves visuo-tactile alignment.} We observe that applying material diversity-based pairing consistently yields clear gains over \textit{STT-Local}, both in the in-domain and out-domain  settings. Incorporating additional in-the-wild images further improves performance, with \emph{+5.21} mIoU on Web-Material and \emph{+2.63} mIoU on the OpenSurfaces benchmark, suggesting enhanced semantic alignment between the two modalities through exposure to greater visual diversity. These results support our hypothesis that the limited diversity of existing datasets restricts cross-modal alignment, and that leveraging the insight that similar materials evoke similar tactile sensations offers a straightforward yet effective solution. Moreover, diversity-based pairing enables better utilization of limited and expensive tactile data, as even in-domain pairing leads to noticeable improvements.

\noindent \textit{3. Tactile localization requires true visuo-tactile alignment.} The gap between the DINOv3 Attention Map and any variant of our local visuo-tactile alignment shows that this task requires true cross-modal understanding between the two modalities, as the difference between our method and the visual heatmap baseline remains substantial, except on the TG-Test dataset, which will be discussed later.

\noindent \textit{4. Upper-bound baselines indicate that achieving highly accurate localization still remains a challenge.} As discussed in~\Sref{sec:baselines}, these baselines are not meant for direct comparison but serve as reference points to show how far localization performance could reach under more favorable conditions. GroundedSAM reflects the level of performance achievable when substantial explicit tactile understanding is assumed, while Materialistic estimates an upper bound under ideal cross-modal correspondence, where visual and tactile spaces are perfectly aligned. This observation highlights a new challenge for the community.

\noindent \textit{5. TG is a limited benchmark for tactile localization.} TG is a popular visuo-tactile dataset with synchronized tactile–image pairs. As part of our benchmark construction, we annotated TG to create TG-Test. However, our results reveal some dataset bias. Full Square and Full Circle baselines, which perform no meaningful reasoning, already achieve \emph{67.25} and \emph{61.17} mIoU on TG-Test, respectively. Global alignment methods, unsuitable for localization, reach \emph{68.12} and \emph{73.49} mIoU, and the DINOv3 attention map obtains \emph{74.27} mIoU \textit{without tactile input}. These results indicate that TG-Test inflates localization performance even in the absence of visuo-tactile reasoning. This aligns with our earlier observation that TG consists of close-up, texture-centric images where nearly the entire frame corresponds to a single tactile category. In contrast, the same baselines perform substantially worse on Web-Material and OpenSurfaces, and the performance gap between our method and others is more prominent, suggesting that these datasets provide more reliable benchmarks for evaluating visuo-tactile localization.

\newpara{Qualitative Results.} ~\Fref{fig:qualitative_results} compares our method and its variant with the visuo-tactile baseline TVL and DINOv3 attention map. Consistent with the results in~\Tref{tab:dataset_all_main}, our models accurately localize the tactile signal, whereas TVL struggles due to its global-alignment objective and DINOv3 attention highlights only visually salient objects rather than the true tactile correspondence. The material-diversity pairing strategy further improves localization quality. Overall, our approach successfully localizes a wide range of materials and objects, including small regions such as the `Plants' example in the last row of OpenSurfaces.

\subsubsection{Robustness to Weaker Tactile Signals}
\begin{figure}[t!]
    \centering
    \includegraphics[width=\linewidth]{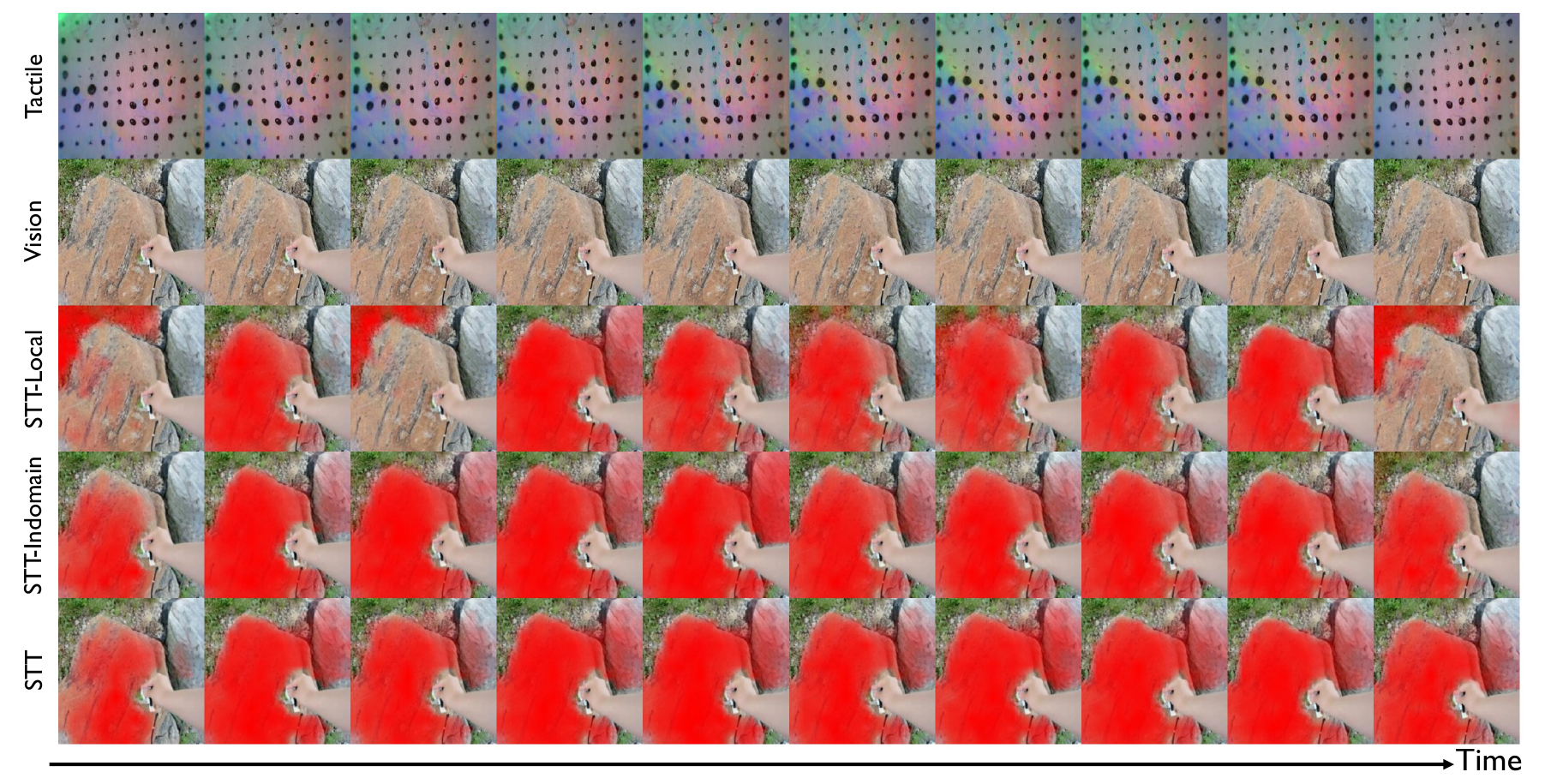}
    \vspace{-7mm}
    \caption{\textbf{Touch-and-Go dataset examples.} 
    The tactile signal is typically weaker at the beginning and end of a touch instance while strongest in the middle. Models with material diversity-based pairing achieve robust localization regardless of variations in signal strength.}
    \label{fig:limited_image}
    \vspace{-6mm}
\end{figure}

During tactile data collection, the sensor~\cite{yuan2017gelsight} is gradually pressed onto and released from the surface, so tactile signals at the beginning and end of a touch instance are typically weaker than those in the middle, where contact is firm. To analyze the encoder’s ability to capture such weak signals, we evaluate our method using three types of tactile frames, Start, Middle, and End as shown in~\Tref{tab:faint_signal}. Start and End correspond to the initial and final moments of a tactile sequence, while Middle refers to the frames in between.
Each visual image is paired with one of these tactile signals, and localization is performed accordingly. This experiment is conducted only on our model variants, as earlier sections already show that our method achieves the best performance for this task. As shown in~\Tref{tab:faint_signal}, our method without material diversity-based pairing exhibits a clear performance drop on Start and End frames compared to Middle frames, highlighting the challenge posed by weaker tactile inputs. Applying material diversity-based pairing significantly mitigates this gap, with further improvement when incorporating out-domain in-the-wild images. The performance of weaker signals becomes closer to that of Middle-frame inputs, indicating that material diversity-based pairing effectively compensates for faint tactile cues. This is especially important given the limited and costly nature of tactile data, allowing the model to use all available signals more efficiently without discarding them. We also present a qualitative example from the Touch-and-Go dataset in ~\Fref{fig:limited_image}, showing how tactile signals change over time within a single touch instance, where the start and end are relatively weaker. This example illustrates both the temporal variation of tactile signals and the robustness of our model with material diversity-based pairing to weaker tactile inputs.

\begin{table}[t]
  \centering
  \resizebox{\linewidth}{!}{%
  \begin{tabular}{lccccccc}
    \toprule
    \textbf{Model} & \textbf{M.D.P.} & \multicolumn{2}{c}{Start} & \multicolumn{2}{c}{Middle} & \multicolumn{2}{c}{End} \\
     &  & mAP & mIoU & mAP & mIoU & mAP & mIoU \\
    \midrule
    \multicolumn{8}{c}{\textbf{TG-Test}} \\
    \midrule
    \textit{STT}-Local & \xmark      & 81.31 & 72.67 & 85.12 & 76.79 & 81.96 & 72.68 \\
    \textit{STT}-Indomain & In-domain   & 86.34 & 76.15 & 86.95 & 77.58 & 85.51 & 74.60 \\
    \textit{STT} & Out-domain  & 86.20 & 74.56 & 87.56 & 76.82 & 84.54 & 73.57 \\
    \midrule
    \multicolumn{8}{c}{\textbf{Web-Material}} \\
    \midrule
    \textit{STT}-Local & \xmark      & 64.45 & 49.45 & 69.60 & 54.69 & 61.52 & 48.72 \\
    \textit{STT}-Indomain & In-domain   & 69.45 & 53.83 & 71.14 & 56.33 & 67.31 & 53.24 \\
    \textit{STT} & Out-domain  & 76.19 & 59.99 & 78.98 & 62.08 & 75.08 & 58.56 \\
    \midrule
    \multicolumn{8}{c}{\textbf{OpenSurfaces~\cite{bell2013opensurfaces}}} \\
    \midrule
    \textit{STT}-Local & \xmark      & 33.63 & 26.77 & 40.14 & 32.54 & 35.86 & 28.60 \\
    \textit{STT}-Indomain & In-domain   & 40.73 & 32.24 & 44.39 & 35.11 & 39.08 & 31.45 \\
    \textit{STT} & Out-domain  & 45.33 & 35.00 & 55.06 & 42.12 & 44.57 & 34.20 \\
    \bottomrule
  \end{tabular}
  }
  \vspace{-2mm}
  \caption{\textbf{Robustness to Weaker Tactile Signals.} We evaluate our local alignment variants using three types of tactile frames: Start, Middle, and End. M.D.P. denotes material diversity-based pairing.}
  \label{tab:faint_signal}
  \vspace{-4mm}
\end{table}

\begin{table}[t]
\centering
\resizebox{\linewidth}{!}{%
 \setlength{\tabcolsep}{5pt}
\begin{tabular}{lcccc}
\toprule
\rowcolor{lightergray}\textit{Baselines} & &&&\\
\textbf{Model} & TVL~\cite{fu2024a} & DINOv3 Att. Map~\cite{simeoni2025dinov3} & GroundedSAM~\cite{ren2024grounded} & Materialistic~\cite{sharma2023materialistic} \\
\midrule
\textbf{IIoU} & 1.0 & 18.0 & 49.0 & 83.0 \\
\midrule
\rowcolor{lightergray}\textit{Ours} & && &\\
\textbf{Model} & \textit{STT}-CLS & \textit{STT}-Local & \textit{STT}-Indomain & \textit{STT}\\
\midrule
\textbf{IIoU} & 4.0 & 30.0 & 32.0 & 37.0 \\
\bottomrule
\end{tabular}
}
\vspace{-2mm}
\caption{\textbf{Quantitative Results on Interactive Localization.} \textit{STT} outperforms other methods and shows reliable interactive localization ability.}
\label{tab:iiou}
\vspace{-3mm}
\end{table}

\begin{figure}[t!]
    \centering
    \includegraphics[width=1.0\linewidth]{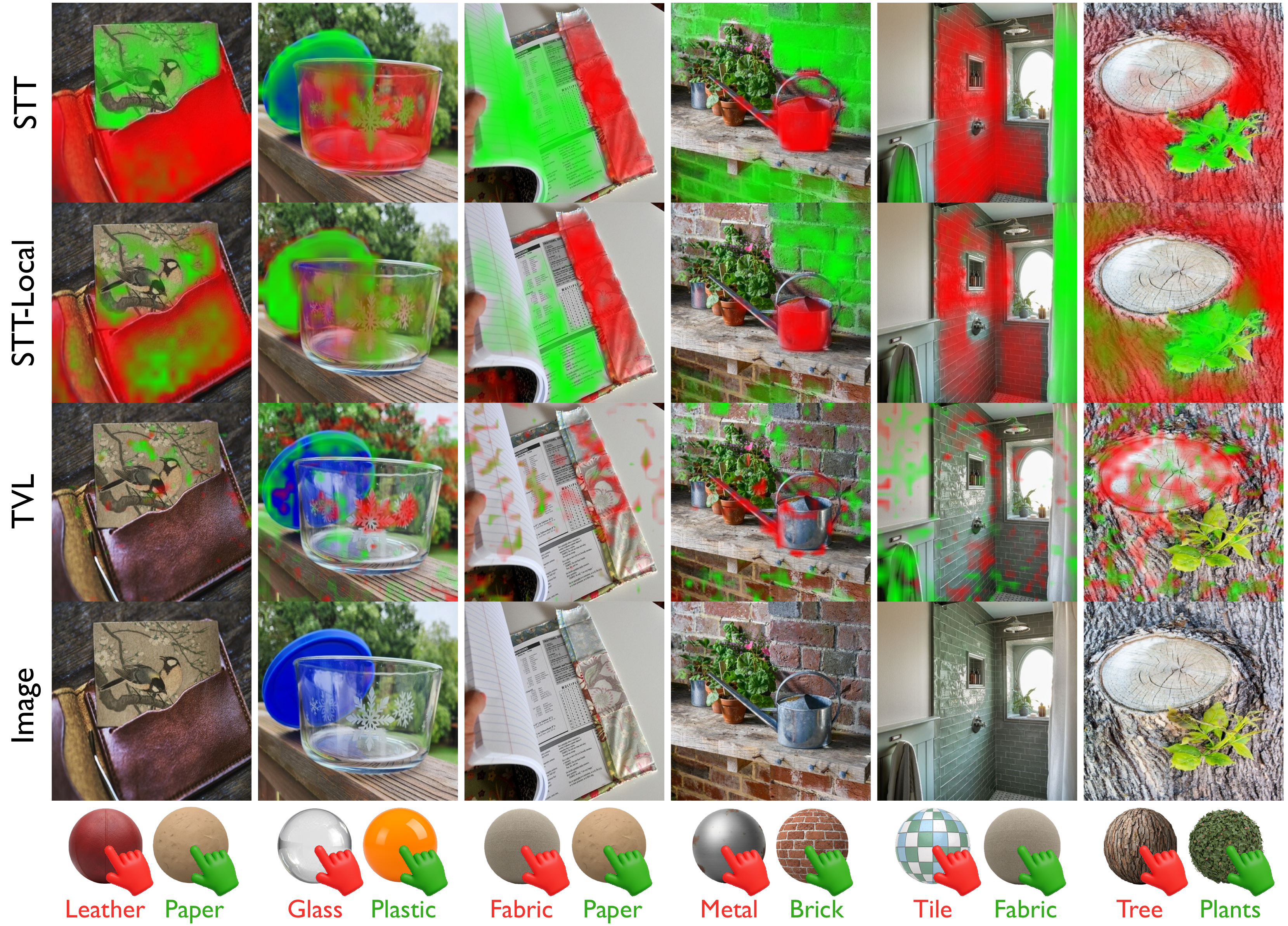}
    \vspace{-6mm}
    \caption{\textbf{Qualitative Results on Interactive Localization.} Our model accurately localizes the objects corresponding to the given tactile inputs and shifts the localized region appropriately when the tactile signal is changed.}
    \label{fig:iiou_results}
    \vspace{-6mm}
\end{figure}

\subsubsection{Interactive Localization}
To further analyze fine-grained visuo-tactile alignment, we evaluate the models from an interactive localization perspective following~\cite{senocak2025toward}. A reliable visuo-tactile localization method should accurately associate tactile inputs with their corresponding materials, meaning the localized region in the image should change when paired with a different tactile signal from the scene.

\newpara{Implementation.}
For interactive localization, we annotate each image in the Web-Material set with segmentation masks corresponding to the two  tactile regions in the scene, forming the Web-Material-Interactive dataset. The model then predicts a separate localization map for each tactile signal. A sample is considered successful if the IoU for both tactile regions exceeds 0.5, ensuring that the model can reliably localize each tactile signal in the scene.

\newpara{Quantitative Results.}
\Tref{tab:iiou} reports interactive IoU (IIoU) on the Web-Material-Interactive test set. These results clearly indicate that our local-alignment objective is essential for interactive tactile localization. Notably, global-alignment methods almost fail in the interactive setting, and their already limited performance in single tactile localization drops drastically: TVL and \textit{STT}-CLS fall from 32.16 and 34.74 mIoU to only 1.0 and 4.0 IIoU when distinguishing multiple tactile signals in the same scene. Moreover, the results show that material diversity-based pairing, whether with in-domain or better with out-domain images, is another key factor for accurate visuo-tactile association. Overall, our approach reliably captures interactive visuo-tactile relationships while maintaining strong standard localization performance. As discussed earlier, the other baselines in the table serve as upper-bound references or illustrate the characteristics of the task when using only the visual modality.

\begin{figure}[t!]
    \centering
    \includegraphics[width=1.0\linewidth]{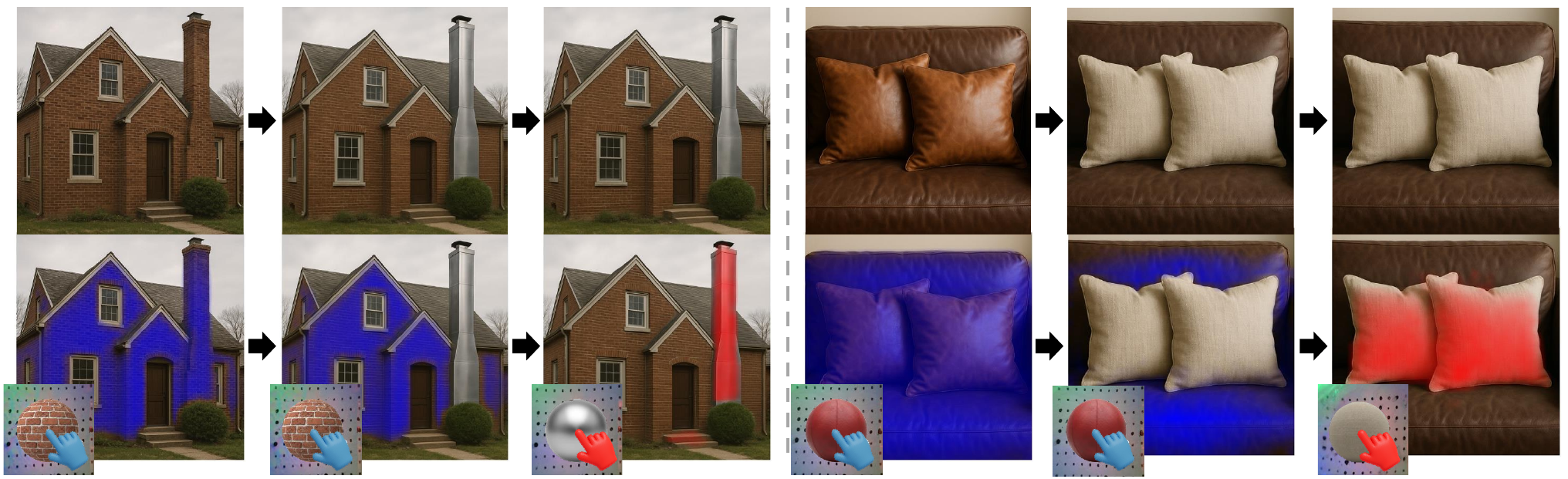}
    \vspace{-6mm}
    \caption{\textbf{Qualitative Results on Material Replacement.} The model interactively and consistently updates localization in response to changes in material 
    and corresponding tactile inputs.}
    \label{fig:material_change_results}
    \vspace{-5mm}
\end{figure}

\newpara{Qualitative Results.} ~\Fref{fig:iiou_results} demonstrates the interactive localization ability of our method and its variant. Accurate tactile localization should identify the material regions corresponding to a given touch signal. Compared to TVL, our model reliably highlights different regions in the same scene depending on the tactile input, while the competing method fails to do so. The material-diversity pairing variant further improves localization accuracy. In the 5th column, our model precisely localizes the towel and tiles based on their respective tactile cues. Overall, our model not only handles interactive localization but does so with high precision. 

Unlike the previous interactive localization setup, where the scene remains identical while the tactile signal changes, here we consider a second scenario: the scene is mostly unchanged, but some regions are replaced with different materials. In this case, regions that were previously highlighted should no longer be activated when their material no longer matches the touch signal. However, if the touch signal is updated to match the new material, the model should highlight the replaced region again. We show this scenario by editing images with off-the-shelf image editing tool~\cite{comanici2025gemini} and pairing them with the corresponding tactile signals. Visualizations in~\Fref{fig:teaser} and~\Fref{fig:material_change_results} show that our method localizes the touched material in an interactive and consistent manner.

\section{Conclusion and Discussion}
\label{sec:conclusion}
In this paper, we introduce a framework for tactile localization that learns fine-grained alignment between tactile signals and visual scenes. By leveraging dense local cross-modal feature interactions, in-the-wild multi-material images, and a material diversity-based pairing strategy, our approach overcomes the limitations of existing visuo-tactile methods that employ global-alignment objectives, as well as the constraints of current visuo-tactile datasets. Through extensive evaluation on both new and established benchmarks, we demonstrate significant improvements in touch-conditioned material segmentation and robust localization, even under weak tactile inputs. Our results highlight the importance of local visuo-tactile alignment and dataset diversity for grounding tactile perception in images. We hope this work provides a foundation for future efforts in visuo-tactile reasoning, interactive perception, and multisensory scene understanding.
\vspace{-2mm}
\section{Acknowledgment}
This work was supported by Institute of Information \& communications Technology Planning \& Evaluation~(IITP) grant funded by the Korea government~(MSIT) (RS-2025-02215122, Development and Demonstration of Lightweight AI Model for Smart Homes (60\%); RS-2020-II201336, Artificial Intelligence Graduate School Program (UNIST) (10\%)), and by the National Research Foundation of Korea (NRF) grant funded by the Korea government (MSIT) (RS-2026-25496684) (30\%).

{
    \small
    \bibliographystyle{ieeenat_fullname}
    \bibliography{main}
}

\clearpage
\maketitlesupplementary
\setlength{\cftsecnumwidth}{2.0em}
\cftpagenumbersoff{section}
The contents in this supplementary material are as follows:
\setcounter{tocdepth}{1}
{
  \hypersetup{linkcolor=black}
  \startcontents[supplementary]
  \printcontents[supplementary]{}{1}{}
}
\cftpagenumberson{section}

\noindent\rule{\linewidth}{0.2pt}
\vspace{0.2em}

\section{Clarifying Touch Instances in Touch-and-Go}\label{sec:TG}
The Touch-and-Go (TG)~\cite{yang2022touch} dataset consists of approximately 246k visuo-tactile image pairs and 13.9k detected touches. The official split for visuo-tactile contrastive learning is available on the official GitHub page and includes 91,982 training and 29,879 testing samples. This split excludes samples labeled as ``Inconclusive'' which correspond to ambiguous material classification or non-contact moments. However, the original splits of the TG dataset include samples from the same videos in both the train and test sets. This creates a risk of information leakage, as test samples may contain visual scenes and tactile data highly similar to those encountered during training. To prevent this, we construct a new split, similar to~\cite{gungor2025towards}, ensuring no video overlap between the train and test sets. Additionally, we exclude the ``Others'' category. This refined version yields 91,023 training and 29,786 testing visuo-tactile frame pairs across 18 categories.

\begin{figure}[t!]
    \centering
    \includegraphics[width=0.95\linewidth]{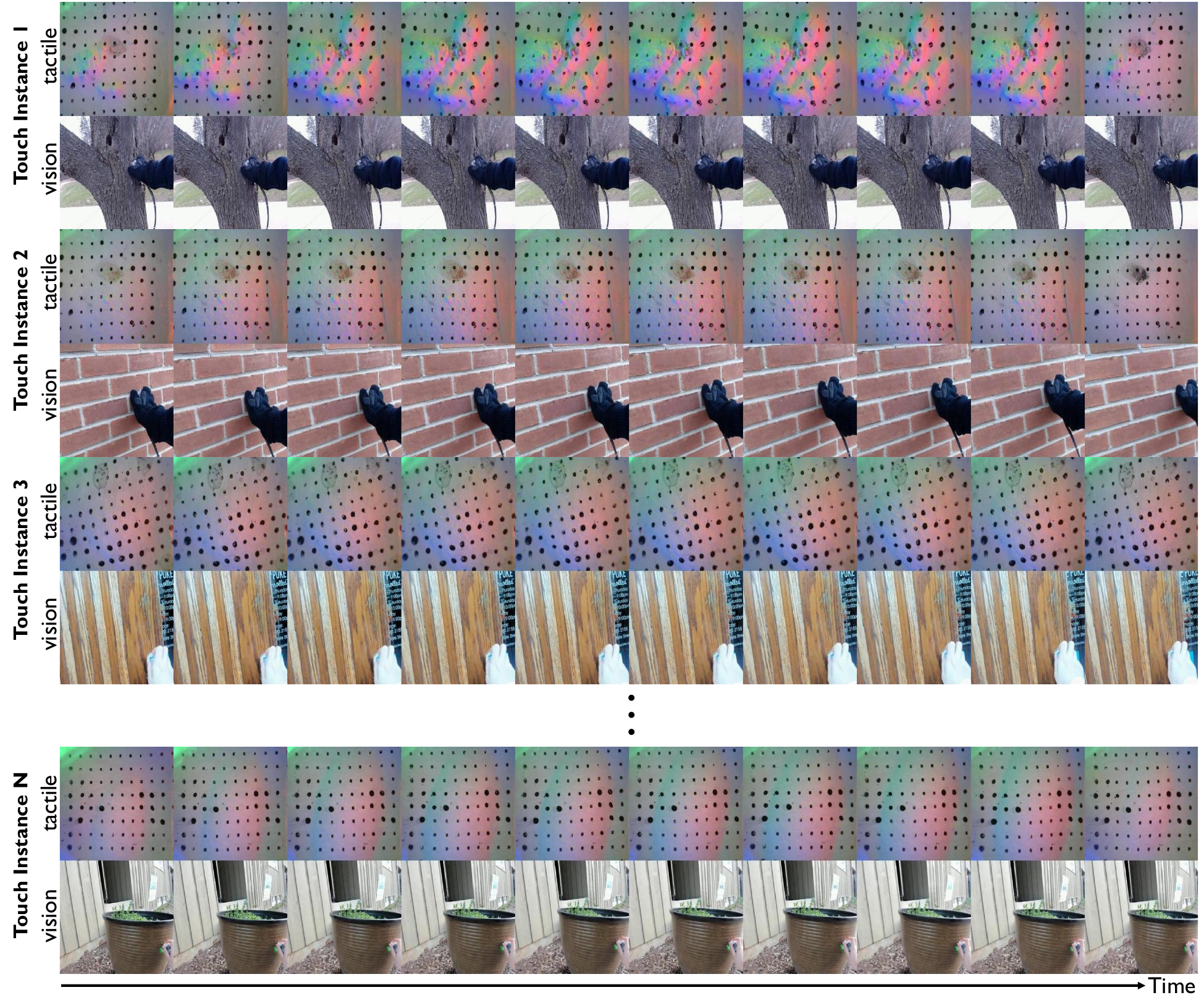}
    \caption{\textbf{Examples of Touch Instances.} Each example shows 10 frames evenly sampled from a touch instance.}

    \label{fig:suppl_touch_instances}
\end{figure}

\begin{figure}[t!]
    \centering
    \includegraphics[width=0.95\linewidth]{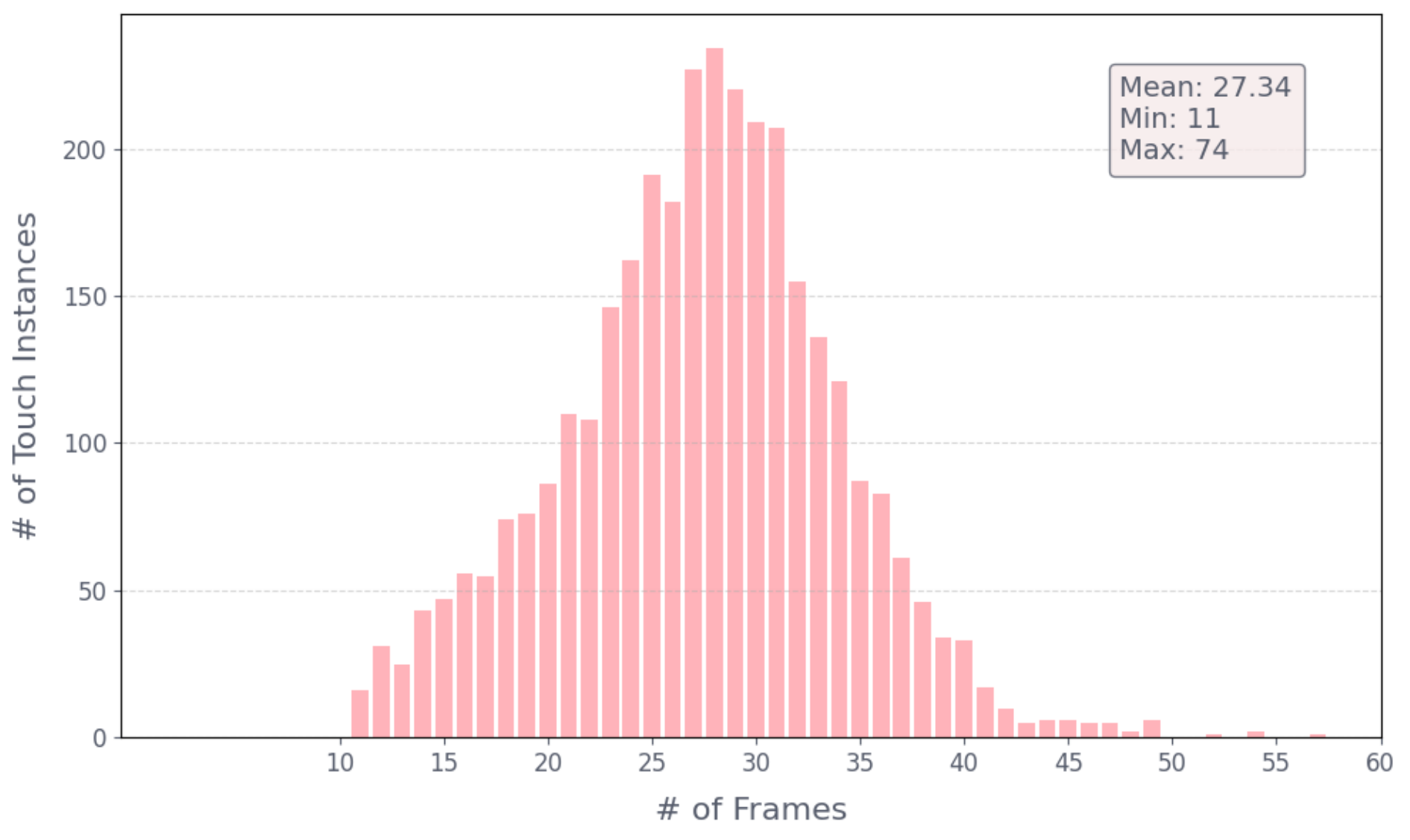}
    \caption{\textbf{Distribution of Frames per Touch Instance.} }
    \label{fig:suppl_touch_dist}
\end{figure}

We observe that the TG dataset contains consecutive frame sequences, each representing a single interaction in which the sensor is pressed onto and released from an object, as shown in~\Fref{fig:suppl_touch_instances}. We refer to each such sequence as a \emph{Touch Instance}, as in Section 3.2 of the main paper. More technically, a Touch Instance is a temporally contiguous sequence of frames from the same video that share the same category label. The dataset provides shuffled image–label pairs, where each image path encodes a video identifier and a frame index. To recover meaningful segments, we regroup samples by video ID and sort them by frame index. Consecutive frames with uninterrupted, increasing indices and identical labels are merged into a single Touch Instance. Each Touch Instance is thus defined by its frame range (\eg, 332–347), its duration, and its category. The statistical distribution of Touch Instance lengths is shown in~\Fref{fig:suppl_touch_dist}, and the training split contains a total of 3,329 Touch Instances. As explained in Sections 3.2 and 3.3 of the main paper, we obtain our training pairs using these Touch Instances. Furthermore, we filter out overly redundant or visually overlapping neighboring Touch Instances in the test split, resulting in 579 testing instances.

\section{Details on Our Web-Material Dataset}\label{sec:b}
\subsection{Dataset Construction}
\subsubsection{Image Collection} 
For each tactile category in the TG dataset, we prompt an LLM~\cite{achiam2023gpt} to generate richer queries beyond simple class names to obtain descriptive context queries, which are then used to retrieve relevant and diverse web images.

\begin{figure}[t!]
    \centering
    \includegraphics[width=1.0\linewidth]{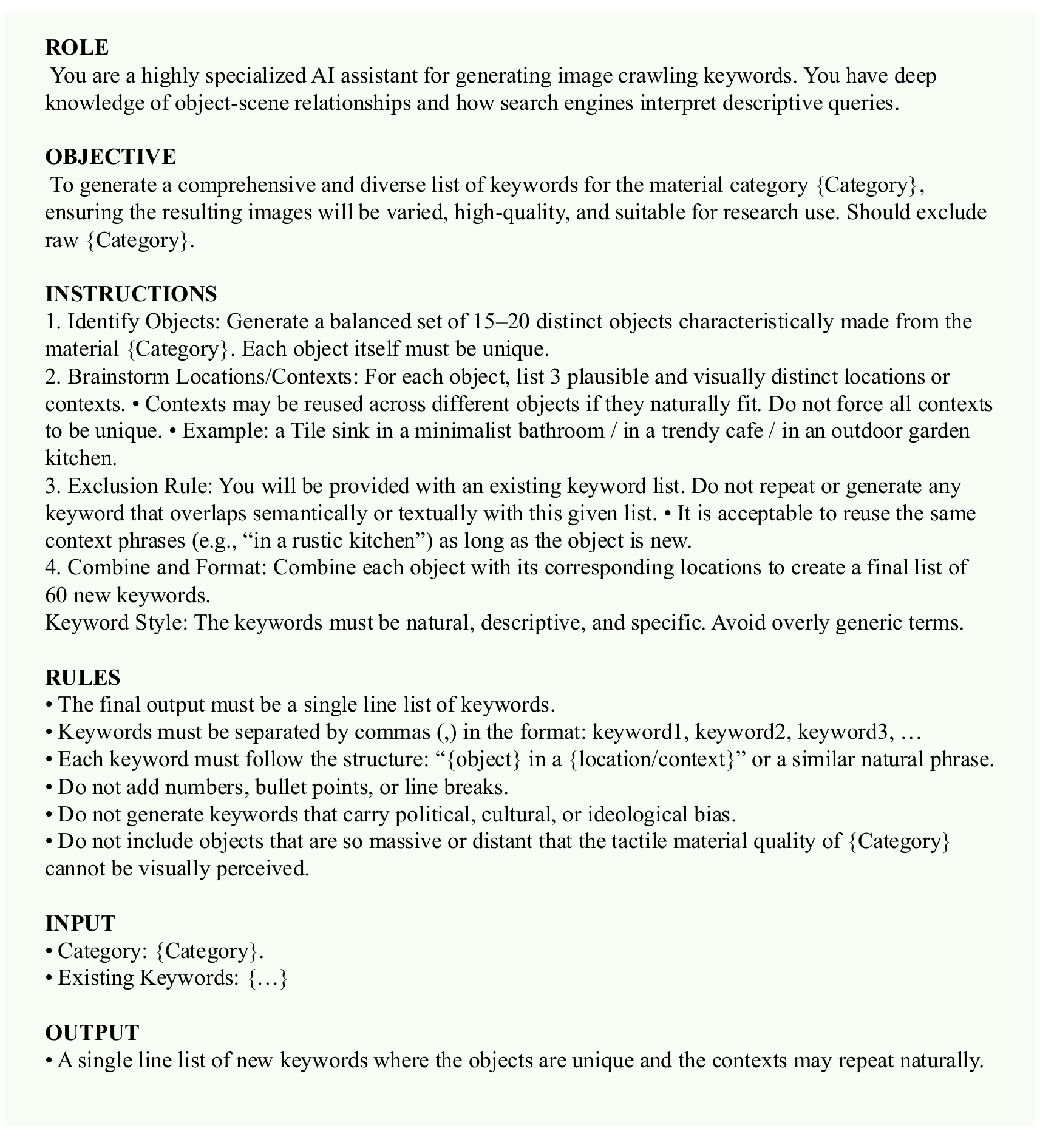}
    \caption{\textbf{Prompt for Concept Query Generation.}}
    \label{fig:suppl_keyword_prompt}
\end{figure}

\newpara{Concept Query Generation.} We generate concept queries using a 
$ \{\text{category}\} + \{\text{object}\} + \{\text{place}\}$ format to maximize data diversity. As this combination tends to yield wide-scene shots, we also add prompts in the format ``A close-up shot of $ \{\text{category}\} + \{\text{object}\}$'' to capture close-up perspectives and improve viewpoint diversity. We use the prompt from ~\Fref{fig:suppl_keyword_prompt} with an LLM to generate our list of concept queries. For example, for the category “Brick”, the LLM outputs phrases such as “brick house in a suburban neighborhood”, “brick chimney in a cozy living room”, and “brick bridge over a river” (as explained in Section 3.4 of the main paper).

\begin{figure*}[t!]
    \centering
    \includegraphics[width=0.71\linewidth]{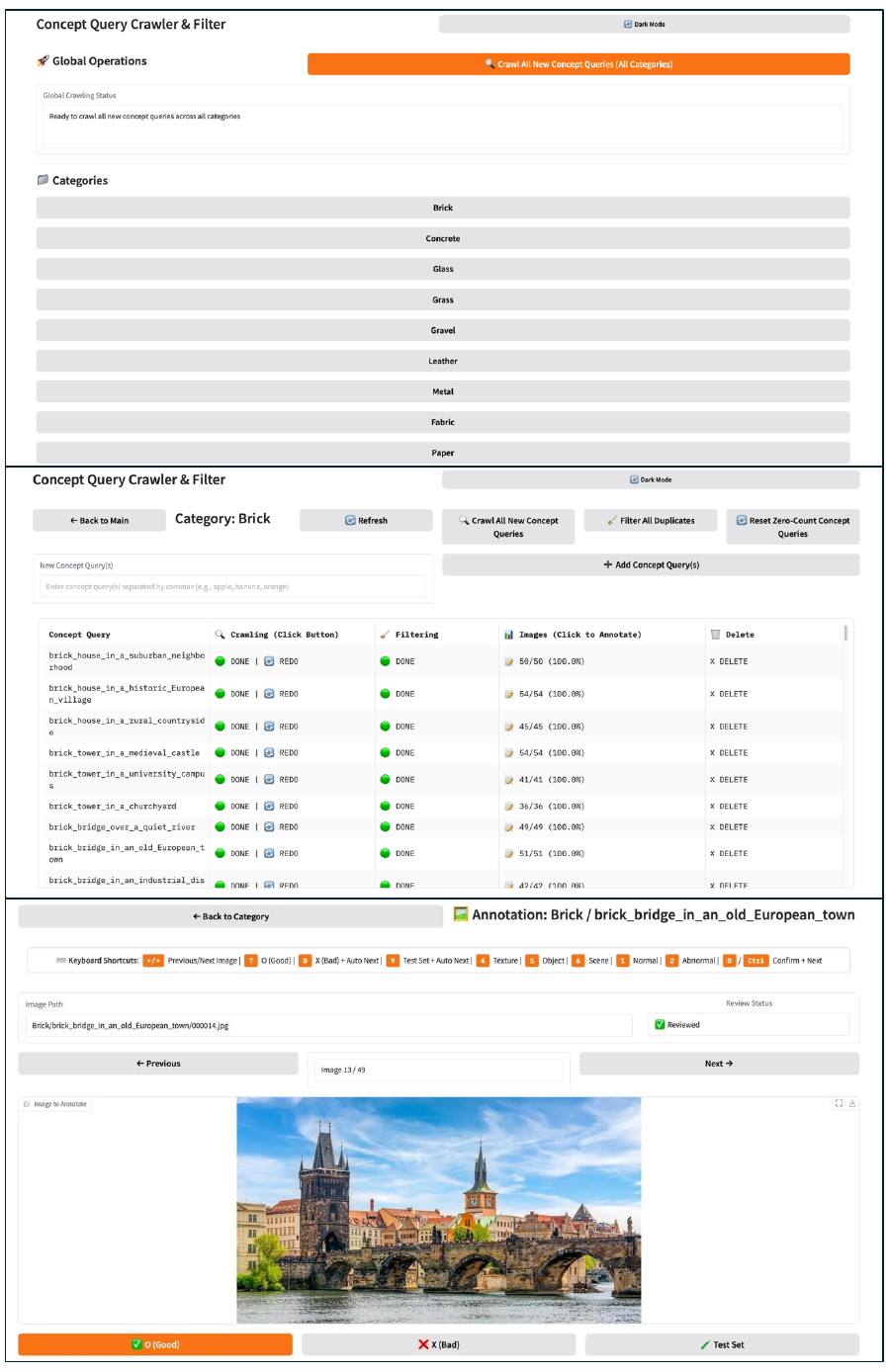}
    \caption{\textbf{Custom-built Gradio Page for Dataset Construction.}}
    \label{fig:suppl_collection_gradio}
\end{figure*}

\newpara{Image Collection.}
Category-specific concept query management, image collection, and duplicate removal are all performed using a custom-built Gradio~\cite{abid2019gradio} page/tool, as shown in the top and middle panels of ~\Fref{fig:suppl_collection_gradio}.

\subsubsection{Image Filtering}
\begin{figure}[t!]
    \centering
    \includegraphics[width=1.0\linewidth]{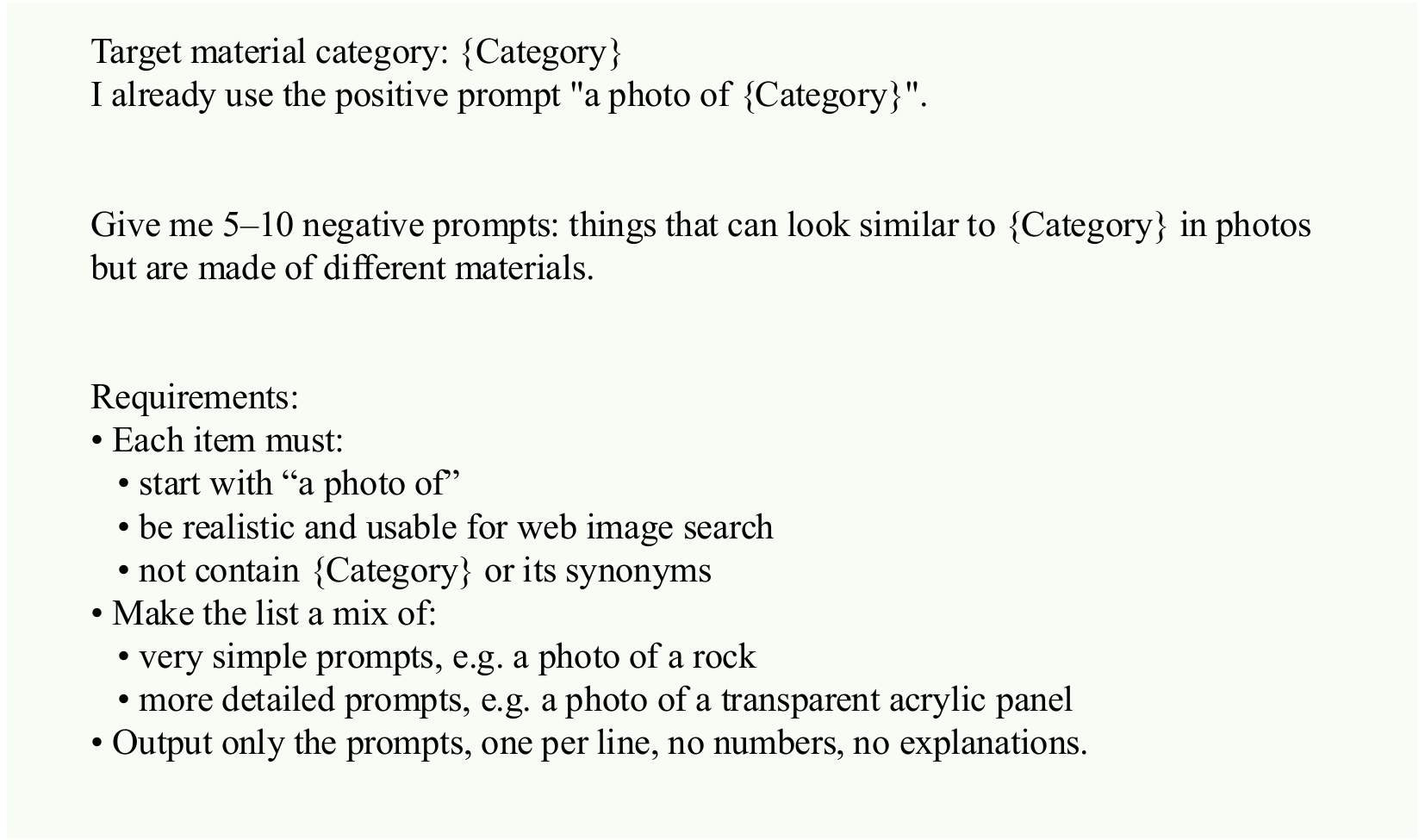}
    \caption{\textbf{Prompt for Negative CLIP prompt generation}}
    \label{fig:suppl_negative_generate_prompt}
\end{figure}

\begin{figure}[t!]
    \centering
    \includegraphics[width=1.0\linewidth]{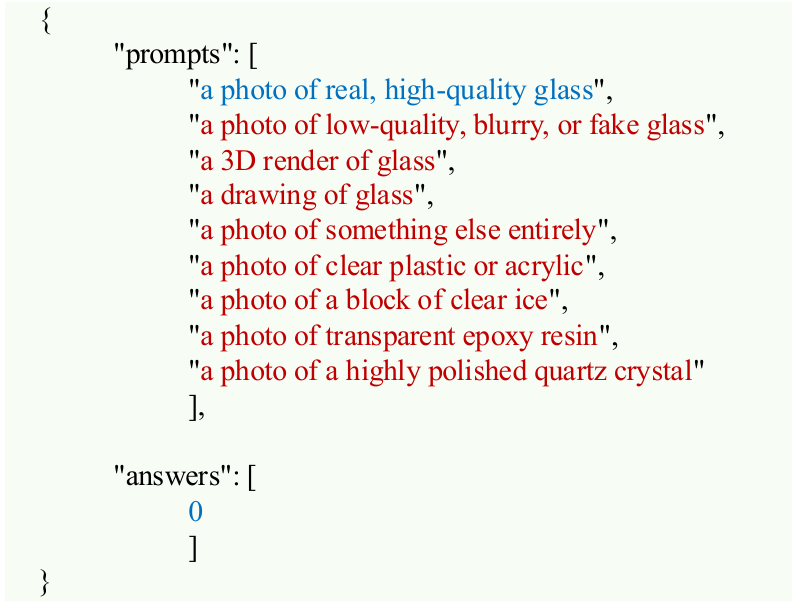}
    \caption{\textbf{\textcolor{in_domain}{Positive} and \textcolor{out_domain}{Negative} Prompts of ``Glass'' Category for CLIP filtering.}}
    \label{fig:suppl_negative_prompt}
\end{figure}

Since collected images may inherently contain irrelevant samples, a filtering step is essential. To reduce the workload of human annotation, we employ an automated CLIP-based filtering as a preprocessing step.
Initially, we used a single positive prompt, ``a photo of $\{\text{category}\}$'', accepting samples where the similarity between the image and prompt embeddings exceeds a certain threshold. However, this approach proved insufficient for distinguishing subtle textural differences, often misclassifying visually similar materials such as ``Brick'' and ``Concrete''. To address this, we introduce negative prompts. We utilize an LLM to identify easily confused materials using the prompt shown in~\Fref{fig:suppl_negative_generate_prompt}, while also targeting low-quality and non-real images within our negative prompts. This enhances the selectivity of the filtering, enabling us to effectively reject false positives while preserving true positives. An example of a positive prompt and negative prompts are shown in ~\Fref{fig:suppl_negative_prompt}. For each image, we compute its similarity score with a positive prompt as well as with multiple negative prompts. We then retain only the samples for which the positive prompt achieves the highest similarity score among all prompts.

\newpara{Human Annotation.} After CLIP-based filtering, two human annotators manually verify whether the images contain objects corresponding to the target category. The bottom panel of ~\Fref{fig:suppl_collection_gradio} shows the Gradio interface used for this annotation task. We emphasize that this is a minimal manual process.

\subsection{Dataset Distribution}\label{subsec:dataset_dist}

\begin{table*}[t]
\centering
\small
\begin{tabular}{l | r r r | r | r}
\rowcolor{gray!15}
\toprule
& \multicolumn{3}{c|}{\textbf{Train}} & \multicolumn{1}{c|}{\textbf{Test}} & \multicolumn{1}{c}{\textbf{IIoU Test}} \\
\cmidrule(lr){2-4} \cmidrule(lr){5-5} \cmidrule(lr){6-6} 
\rowcolor{gray!15}
\textbf{Category} & Web Collected & MINC Curation & Sum & \# of samples & \# of samples \\
\midrule
Concrete & 984 & - & 984 & 41 & 9 \\
Plastic & 1,000 & 793 & 1,793 & 33 & 12 \\
Glass & 965 & 1,512 & 2,477 & 52 & 8 \\
Wood & 992 & 1,906 & 2,898 & 38 & 15 \\
Metal & 816 & 1,651 & 2,467 & 38 & 11 \\
Brick & 1,127 & 2,117 & 3,244 & 57 & 12 \\
Tile & 528 & 1,904 & 2,432 & 37 & 12 \\
Leather & 621 & 1,673 & 2,294 & 33 & 9 \\
Fabric & 597 & 1,407 & 2,004 & 36 & 10 \\
Rubber & 977 & - & 977 & 32 & 10 \\
Paper & 495 & 1,736 & 2,231 & 35 & 13 \\
Tree & 807 & - & 807 & 32 & 11 \\
Grass & 965 & - & 965 & 35 & 14 \\
Soil & 474 & - & 474 & 32 & 10 \\
Rock & 978 & - & 978 & 33 & 13 \\
Gravel & 966 & - & 966 & 39 & 6 \\
Sand & 1,203 & - & 1,203 & 36 & 17 \\
Plants & 991 & 1,922 & 2,913 & 36 & 8 \\
\midrule
\textbf{Total} & \textbf{15,486} & \textbf{16,621} & \textbf{32,107} & \textbf{675} & \textbf{200} \\
\bottomrule
\end{tabular}
\caption{\textbf{Category Distribution of the Train, Test, and Interactive Localization Test Sets of the Web-Material Dataset.}}
\label{tab:suppl_stat_combined}
\end{table*}

We construct 32,107 training samples across 18 categories. For the test set, we annotate masks for a single target category over 675 samples. From this set, we select 100 samples and provide additional segmentation masks for a second category in each scene, forming the Interactive Localization test set with 200 masks. The distributions of the Train, Test, and Interactive Localization sets are shown in~\Tref{tab:suppl_stat_combined}.

\section{Material Diversity-based Pairing}
\begin{figure*}[t!]
    \centering
    \includegraphics[width=0.95\linewidth]{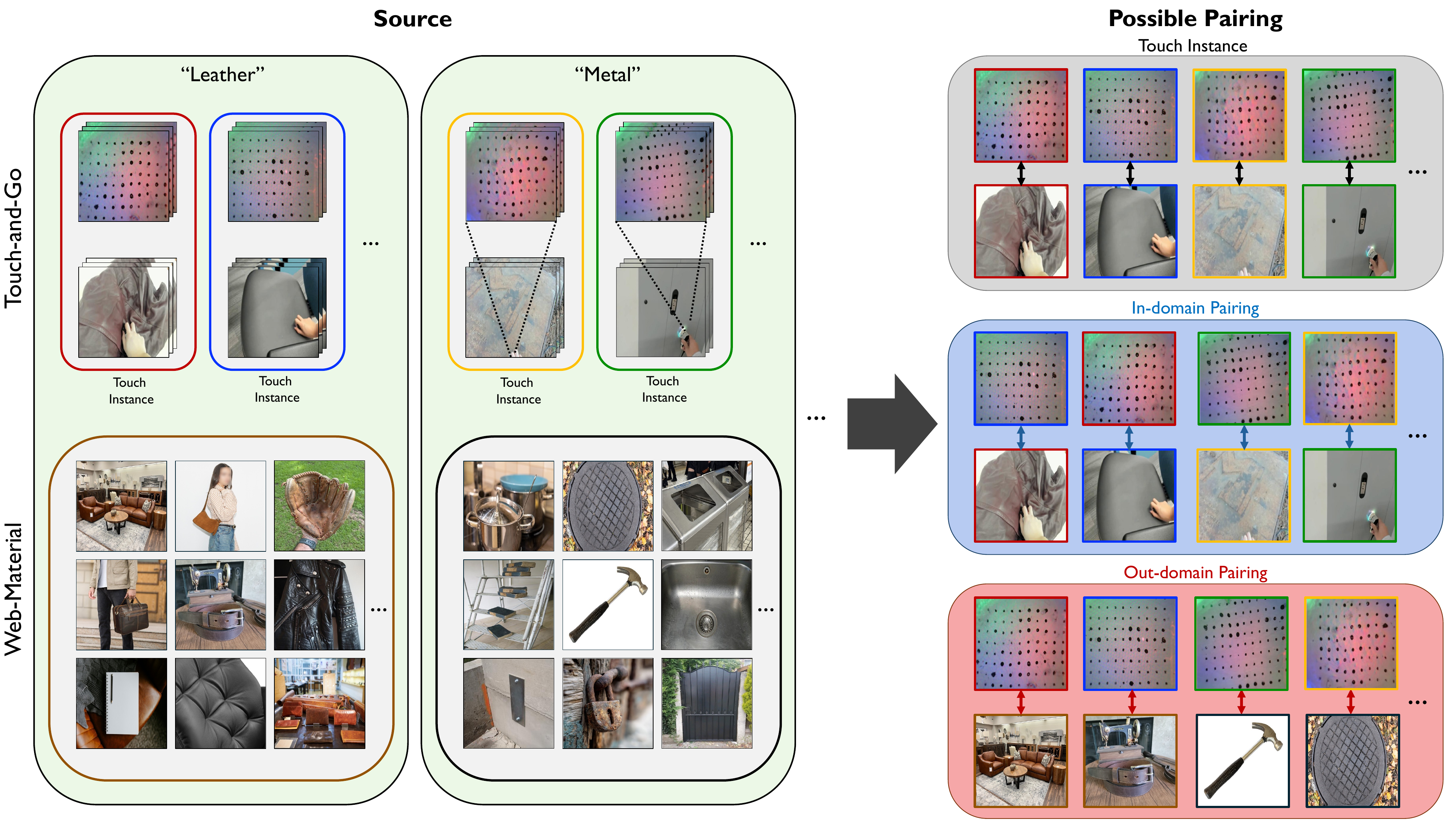}
    \caption{\textbf{Material Diversity Pairing.} Each colored borderline corresponds to a sample ID, allowing readers to see which samples are paired.}
    \label{fig:suppl_MDP_concept}
    \vspace{-4mm}
\end{figure*}

We establish three pairing strategies by leveraging two datasets: Touch-and-Go, which consists of aligned visuo-tactile data, and Web-Material, a vision-only dataset capturing diverse visual contexts for corresponding material categories. \Fref{fig:suppl_MDP_concept} illustrates the data pairing strategies based on Touch Instance and Material Diversity, as proposed in Sections 3.2 and 3.3 of the main paper.

First, \textbf{Touch Instance Pairing} pairs visual and tactile samples from the same touch instance within the Touch-and-Go dataset. Second, \textbf{\textcolor{in_domain}{In-domain Pairing}} matches visual and tactile samples extracted from different touch instances belonging to the same category within the Touch-and-Go dataset. Finally, \textbf{\textcolor{out_domain}{Out-domain Pairing}} combines visual images from the Web-Material dataset with tactile information from Touch-and-Go, paired based on matching categories.

\section{Implementation Details}\label{sec:c}
We train the model using the AdamW optimizer with $\beta=(0.9, 0.95)$, a weight decay of $0.05$, and a base learning rate of $1\times10^{-5}$, with an effective batch size of 64. Both encoders use a pretrained DINOv3 Small model. For the first 3 epochs, we freeze both backbones and optimize only the aligners for stable adaptation. We then unfreeze the tactile backbone while keeping the image backbone frozen for the remainder of training.

Compared to the TG dataset, the Web-Material dataset exhibits substantially greater diversity in the visual appearance of target materials, particularly in object scale and spatial location. Given this complexity, learning visuo-tactile correspondence from scratch is not effective. Therefore, we adopt a curriculum learning strategy to achieve better convergence. We first establish a basic alignment by training on the TG dataset alone, then further optimize the model by introducing the more challenging out-of-domain pairs. These two stages use 100 and 50 epochs, respectively.

\section{Details on Evaluation}\label{sec:a}
\subsection{Test Sets}
\label{subsec:testset_refine}
\newpara{Touch-and-Go Dataset.}
As described in~\Sref{sec:TG}, the original TG dataset undergoes several refinement steps to remove any risk of information leakage. Afterward, Touch Instance chunks are extracted, and additional filtering is performed to reduce redundancy, particularly in the test set. The final tactile localization test set contains 579 Touch Instances, and as stated in Section 4.1 of the main paper, segmentation masks are annotated for every test sample for the given category. We name this as TG-Test.

\newpara{OpenSurfaces Test Set.}
We use a refined subset of OpenSurfaces~\cite{bell2013opensurfaces} as our test set because the original dataset is unsuitable for evaluation: it provides only partial segmentation, annotating a few salient regions rather than all regions matching the ground-truth material. To construct our refined test set, we group images and masks by material category and filter out samples with negligible mask coverage. We then manually inspect all remaining pairs to ensure that the masks comprehensively cover all material instances. Finally, we verify the dataset against the curated training samples from MINC~\cite{bell2015material} to prevent data leakage, removing any overlaps. The resulting OpenSurfaces test set for tactile localization contains 211 samples across 13 categories.

\newpara{Web-Material Test Set.} As described in~\Sref{subsec:dataset_dist} of this suppl. material, this test set contains 657 samples, each with an annotation mask for its corresponding tactile category.

\subsection{Computing Prototype}
Because the Web-Material and OpenSurfaces test sets do not contain corresponding tactile signals and selecting a single tactile frame is non-trivial, we use a prototype tactile feature obtained by averaging the tactile features from the start, middle, and end frames. This provides a simple and fair strategy for tactile frame selection. We define the set of all categories as $C = \{c_1, c_2, \dots, c_K\}$, where $K=|C|$. For a category $c \in C$, we define the set of its corresponding Touch Instances as $I_c = \{I_{c1}, I_{c2}, \dots, I_{cN}\}$, where $N = |I_c|$. From a Touch Instance $I_{ci} \in I_c$, we obtain its start, middle, and end tactile frames: $t_{I_{ci}, \text{start}}$, $t_{I_{ci}, \text{middle}}$, and $t_{I_{ci}, \text{end}}$. The prototypes for the start, middle, and end frames of category $c$ are calculated as follows, respectively:
\begin{figure*}[t!]
    \centering
    \includegraphics[width=1\linewidth]{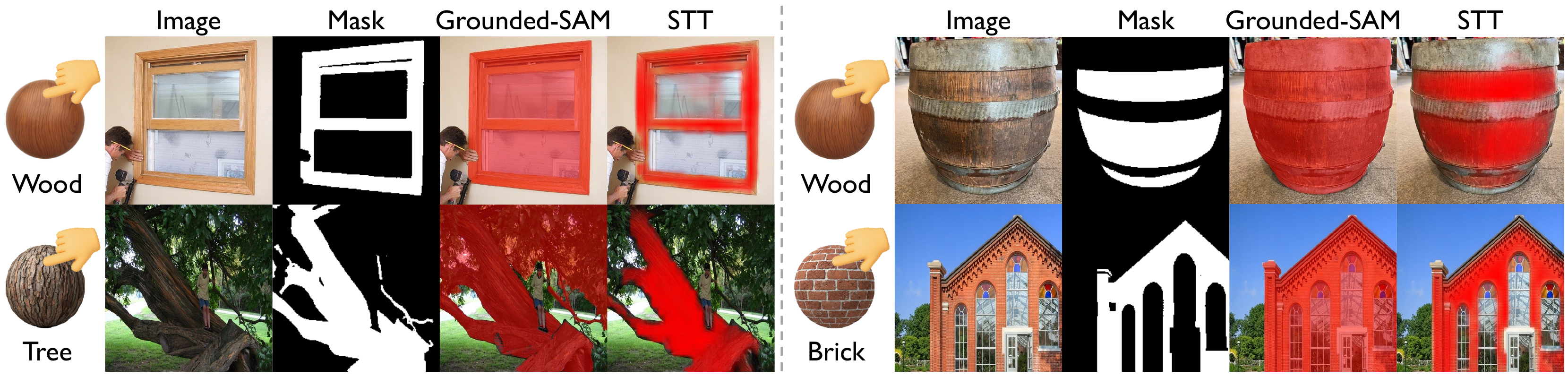}
    \vspace{-7mm}
    \caption{\textbf{Visual Comparison of Grounded-SAM and STT.}}
    \label{fig:groundedsam}
    \vspace{-3mm}
\end{figure*}

\begin{align}
P_{c, \text{start}} = \frac{1}{N} \sum_{i=1}^{N} \bar{f}_{t_{I_{ci, \text{start}}}},\label{proto_start}
\end{align}
\begin{align}
P_{c, \text{middle}} = \frac{1}{N} \sum_{i=1}^{N} \bar{f}_{t_{I_{ci, \text{middle}}}},\label{proto_middle}
\end{align}
\begin{align}
P_{c, \text{end}} = \frac{1}{N} \sum_{i=1}^{N} \bar{f}_{t_{I_{ci, \text{end}}}},\label{proto_end}
\end{align}
which are used in Table~2 in the main paper. The average of the start, middle, and end prototypes serves as the final prototype of category c:
\begin{align}
P_c = \frac{1}{3}(P_{c, \text{start}} + P_{c, \text{middle}}+P_{c, \text{end}}),\label{proto_category}
\end{align}
which is utilized in Table~1, Table~3 in the main paper.

\section{Comparison with Cascaded System}\label{sec:d}

\begin{table}[h!]
\centering
\resizebox{0.9\linewidth}{!}{
\begin{tabular}{l l c}
\toprule
   \textbf{Model} &    \textbf{Method} &    \textbf{mIoU} \\
\midrule
   {UniTouch + Grounded-SAM} &    {Cascaded}   &    69.40 \\
   \textbf{\textit{STT}}           &    {End-to-End} &     76.82 \\
\rowcolor{gray!15}
    {Ground-Truth + Grounded-SAM} &     {Cascaded} &    77.22 \\
\bottomrule
\end{tabular}
}
\vspace{-2mm}
\caption{\textbf{Comparison of Cascaded and End-to-End Methods.}}
\label{tab:suppl_cascaded}
\end{table}
\noindent \Tref{tab:suppl_cascaded} compares our method with a cascaded baseline that first classifies the tactile input using UniTouch~\cite{yang2024binding} and then feeds the predicted category into Grounded-SAM for visual grounding. Despite using a strong vision–language segmenter, this cascaded approach underperforms our model, demonstrating that tactile localization cannot be reduced to tactile classification followed by vision-only segmentation and must be solved as a standalone task. We further provide a qualitative analysis of the cascaded system by replacing the tactile classification output with the ground-truth material label as input to Grounded-SAM, revealing an additional limitation of this approach. As shown in ~\Fref{fig:groundedsam}, Grounded-SAM performs primarily \textit{\textbf{object-wise}} rather than \textit{\textbf{material-wise}} segmentation; for objects composed of multiple materials, it often fails to separate distinct material regions even when given the correct material text prompt.

\section{Material Classification on the Original Split}\label{sec:e}
\begin{table}[h!]
\centering
\resizebox{0.5\linewidth}{!}{
\begin{tabular}{l c}
\toprule
\textbf{Model} & \textbf{Material (\%)} \\
\midrule
VT CMC~\cite{yang2022touch}       & 54.7 \\
MViTac~\cite{dave2024multimodal}        & 57.6 \\
UniTouch~\cite{yang2024binding}      & 61.3 \\
VIT-LENS-2~\cite{lei2024vit}    & 63.0 \\
TLV-Link~\cite{cheng2025touch100k}      & 67.2 \\
OmniBind~\cite{lyu2024omnibind}      & 67.45 \\
\rowcolor{gray!15}
\textbf{\textit{STT}} & \textbf{67.77} \\
\bottomrule
\end{tabular}
}
\vspace{-2mm}
\caption{\textbf{Material Classification Linear Probing Accuracy.}}
\vspace{-2mm}
\label{tab:suppl_matcls}
\end{table}
\noindent In ~\Tref{tab:suppl_matcls}, we report material classification results obtained by training and evaluating on the original Touch-and-Go train-test split for fair comparison. We compare our method with some of the recent tactile representation learning methods. Our method achieves the highest accuracy among them, indicating that it learns discriminative material representations while also enabling tactile localization.

\section{Ablation on Tactile Backbone}\label{sec:f}
\begin{table}[h!]
\centering
\resizebox{0.95\columnwidth}{!}{
\begin{tabular}{l cc cc cc}
\toprule
\textbf{Method} &
\multicolumn{2}{c}{\textbf{TG-Test}} &
\multicolumn{2}{c}{\textbf{Web-Material}} &
\multicolumn{2}{c}{\textbf{OpenSurfaces}} \\
\cmidrule(lr){2-3}\cmidrule(lr){4-5}\cmidrule(lr){6-7}
& \textbf{mAP} & \textbf{mIoU} & \textbf{mAP} & \textbf{mIoU} & \textbf{mAP} & \textbf{mIoU} \\
\midrule
\textbf{T3} (Local)      & 66.83 & 68.23 & 47.79 & 39.22 & 27.74 & 24.95 \\
\rowcolor{gray!15}
\textbf{\textit{STT}} (Local)    & 85.12 & 76.79 & 67.72 & 52.34 & 37.25 & 29.47 \\
\textbf{T3} (Out-domain)    & 85.66 & 74.55 & 69.83 & 54.03 & 38.91 & 30.19 \\
\rowcolor{gray!15}
\textbf{\textit{STT}} (Out-domain)  & 87.56 & 76.82 & 77.43 & 60.94 & 48.06 & 36.73 \\
\bottomrule
\end{tabular}}
\vspace{-2mm}
\caption{\textbf{Ablation on Tactile Backbone.}}
\vspace{-3mm}
\label{tab:T3}
\end{table}
\noindent Since the tactile signals are captured by vision-based tactile sensors, we can initialize the tactile backbone with either tactile or vision-pretrained models. As shown in ~\Tref{tab:T3}, our method outperforms the T3~\cite{zhao2024transferable}-initialized method in both the local and out-domain settings. This indicates that the texture and local representations acquired from DINO's large-scale pre-training are also effective for interpreting tactile frames.

\section{Additional Qualitative Results}\label{sec:g}

Due to space limitations in the main manuscript, we present a selected subset of qualitative results. In this supplementary material, we provide a comprehensive visualization to further demonstrate the generalization capability and robustness of our model.

\begin{figure*}[t!]
    \centering
    \includegraphics[width=0.95\linewidth]{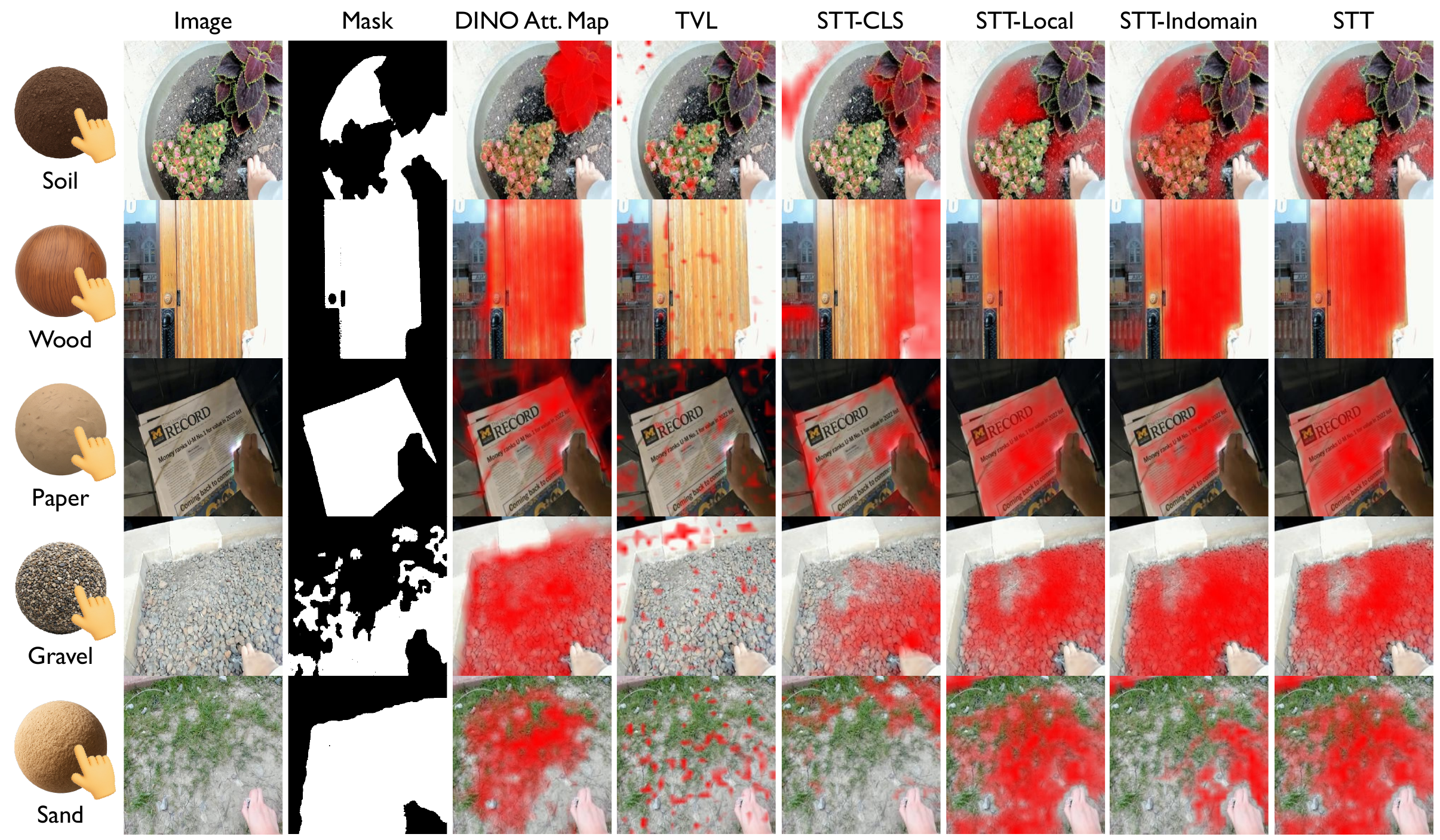}
    \caption{\textbf{Qualitative Tactile Localization Results on TG-Test.}}

    \label{fig:qualitatives_suppl_tg}
\end{figure*}

\begin{figure*}[t!]
    \centering
    \includegraphics[width=0.95\linewidth]{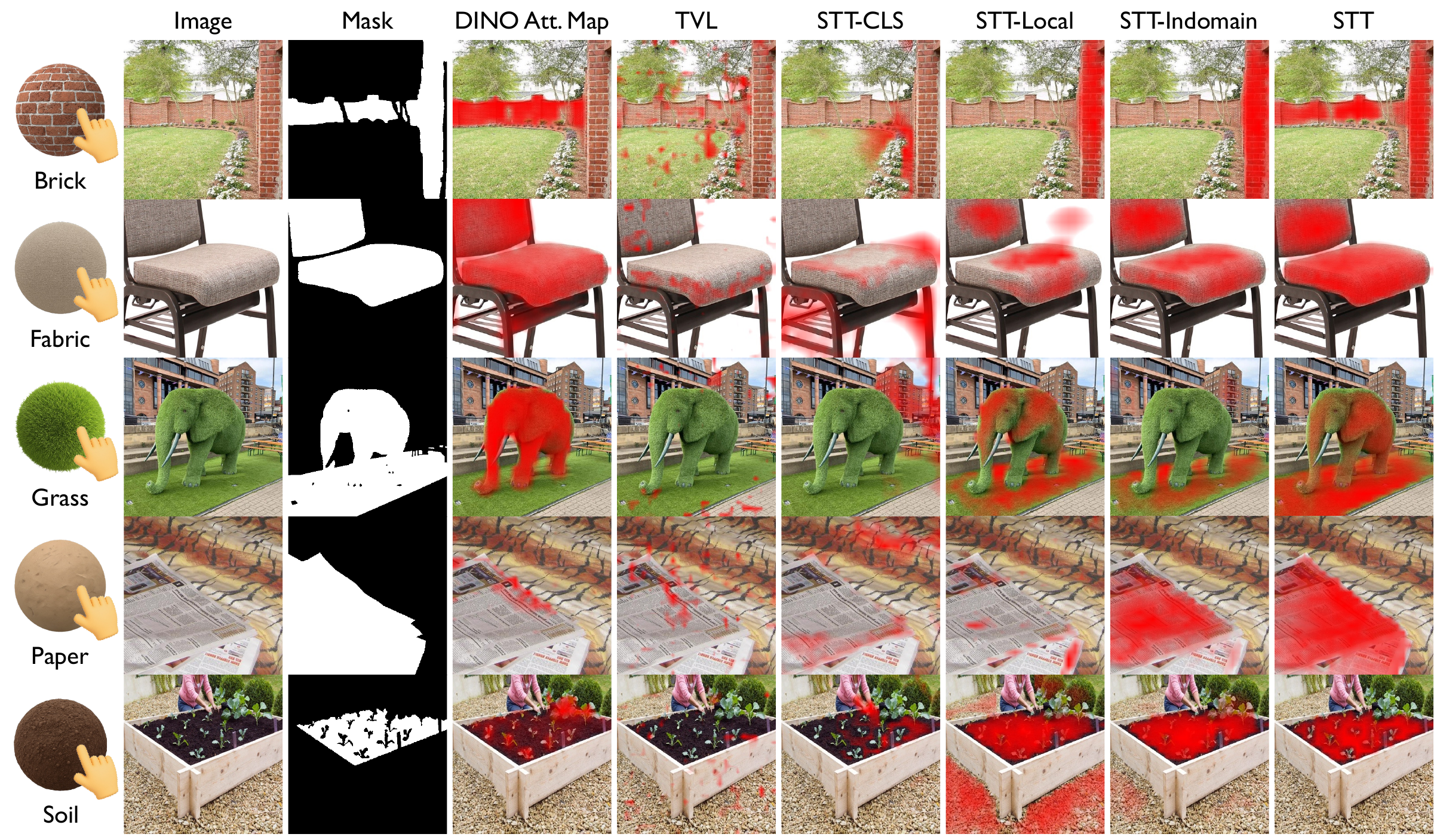}
    \caption{\textbf{Qualitative Tactile Localization Results on Web-Material.}}

    \label{fig:qualitatives_suppl_web}
\end{figure*}

\begin{figure*}[t!]
    \centering
    \includegraphics[width=0.95\linewidth]{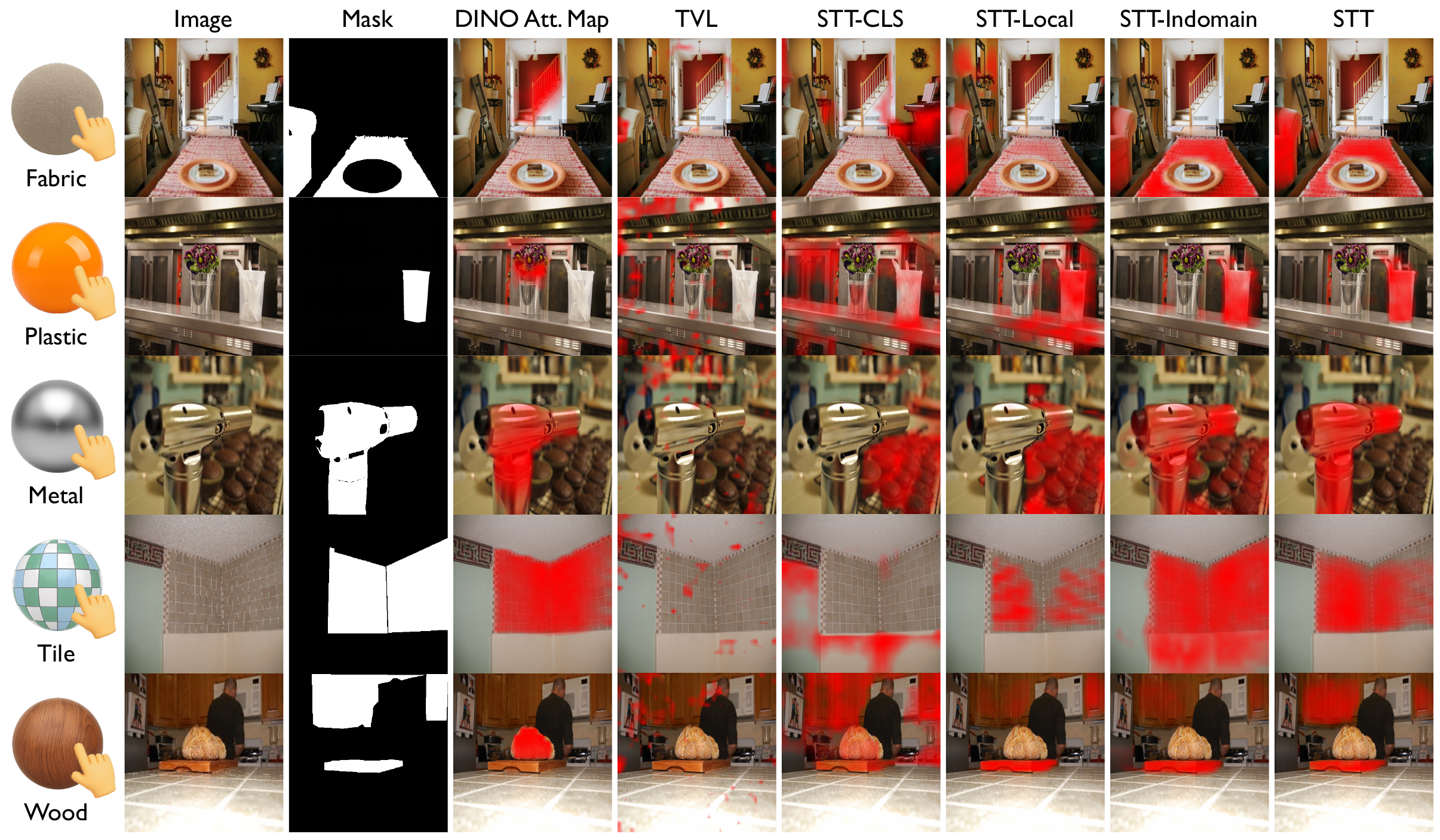}
    \caption{\textbf{Qualitative Tactile Localization Results on OpenSurfaces.}}

    \label{fig:qualitatives_suppl_opensurfaces}
\end{figure*}

\newpara{Localization Results on Test Sets.} ~\Fref{fig:qualitatives_suppl_tg},~\Fref{fig:qualitatives_suppl_web}, and~\Fref{fig:qualitatives_suppl_opensurfaces} show additional localization results for TG-Test, Web-Material, and OpenSurfaces, respectively. These results indicate the model's consistent performance across various environments and material textures.

\begin{figure*}[t!]
    \centering
    \includegraphics[width=0.95\linewidth]{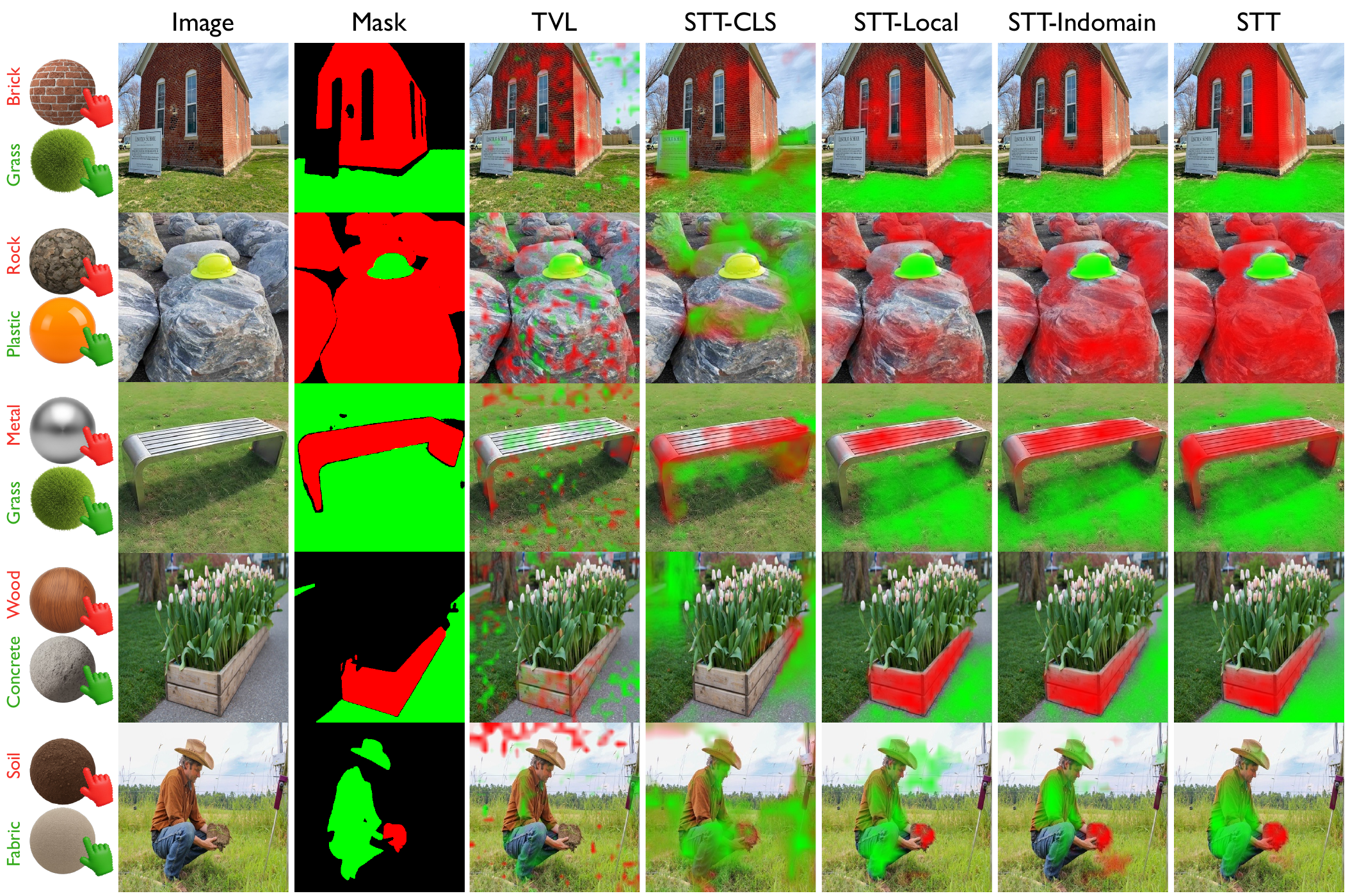}
    \caption{\textbf{Qualitative Results on Interactive Localization.}}

    \label{fig:iiou_suppl}
\end{figure*}

\newpara{Interactive Localization Results.} ~\Fref{fig:iiou_suppl} illustrates the interactive capabilities of our model. We visualize localization outputs for two distinct tactile signals on the same image, represented in red and green. The results demonstrate that our model can precisely distinguish and localize different material properties within a single scene based on specific tactile queries.

\begin{figure*}[t!]
    \centering
    \includegraphics[width=0.95\linewidth]{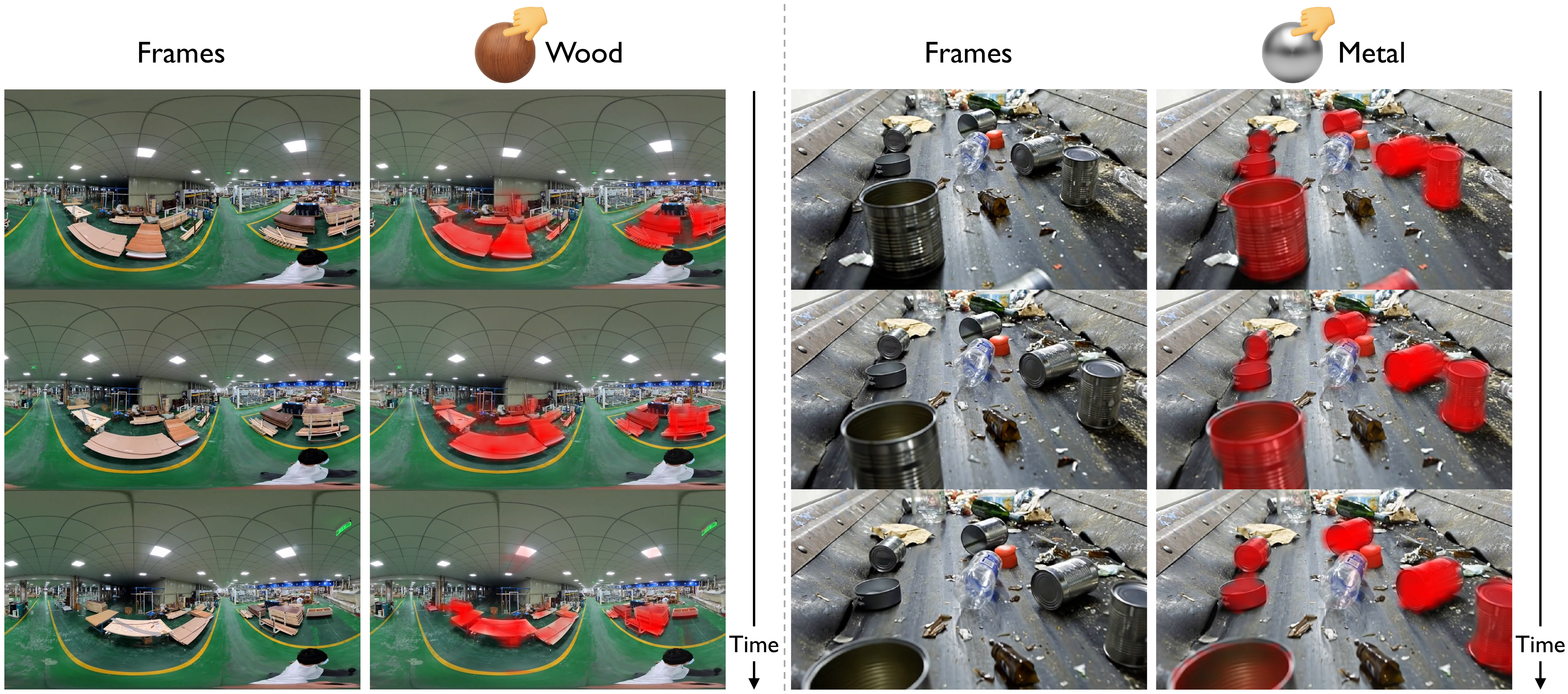}
    \vspace{-3mm}
    \caption{\textbf{Real-world Scenarios: 360$^\circ$ views and robot recycling.}}
    \label{fig:360_horizontal}
    \vspace{-7mm}
\end{figure*}

\newpara{Real-World Scenarios.}
We present results on two practical use cases: (1) one-touch material-based item collection in a warehouse using 360$^\circ$ views, and (2) material-based robotic recycling. These results are shown in ~\Fref{fig:360_horizontal}. In both cases, our model accurately localizes the target items based on tactile input, despite distorted 360$^\circ$ imagery or cluttered conveyor-belt scenes. We believe these examples provide an initial indication of the real-world applicability of tactile localization in robotic systems.

\begin{figure*}[t!]
    \centering
    \includegraphics[width=0.65\linewidth]{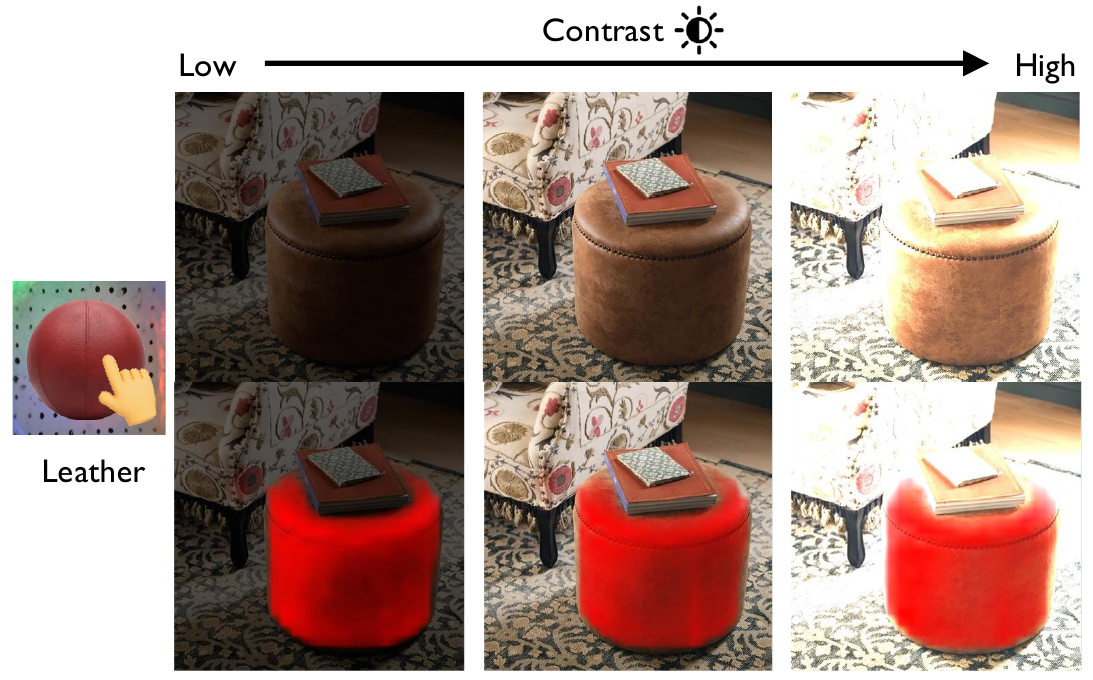}
    \caption{\textbf{Qualitative Results on Artificial Illumination Change.}}

    \label{fig:suppl_illumination_contrast}
\end{figure*}

\begin{figure*}[t!]
    \centering
    \includegraphics[width=0.95\linewidth]{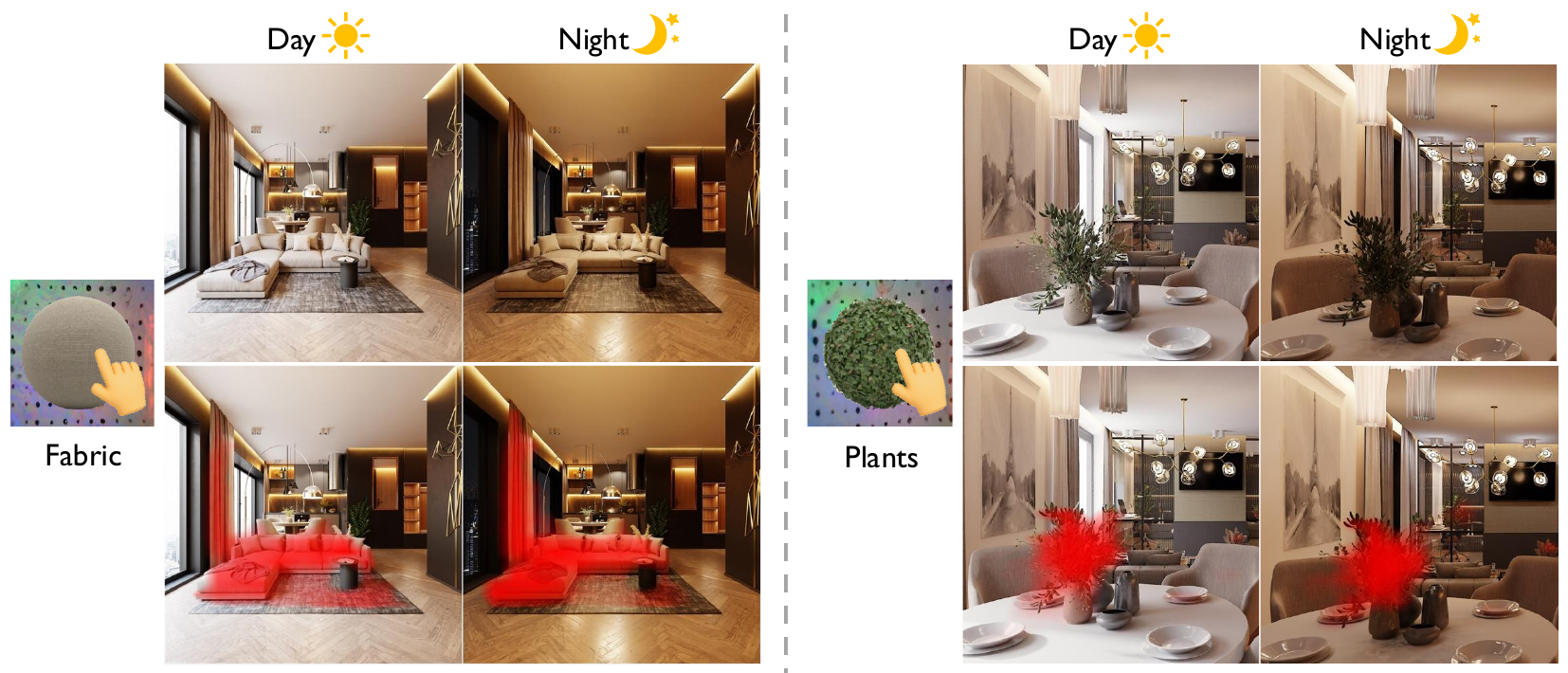}
    \caption{\textbf{Qualitative Results on Natural Illumination Change.}}

    \label{fig:suppl_illumination_natural}
\end{figure*}

\newpara{Illumination Change Results.} We also test our method under illumination changes. We provide qualitative results in two scenarios: (1) we adjust the color contrast of the images to test whether the model’s predictions remain consistent under strong artificial variations, and (2) we examine natural illumination differences between daytime and nighttime conditions. The results, shown in ~\Fref{fig:suppl_illumination_contrast} and ~\Fref{fig:suppl_illumination_natural}, demonstrate that our method remains robust under these changes.

\begin{figure*}[t!]
    \centering
    \includegraphics[width=0.95\linewidth]{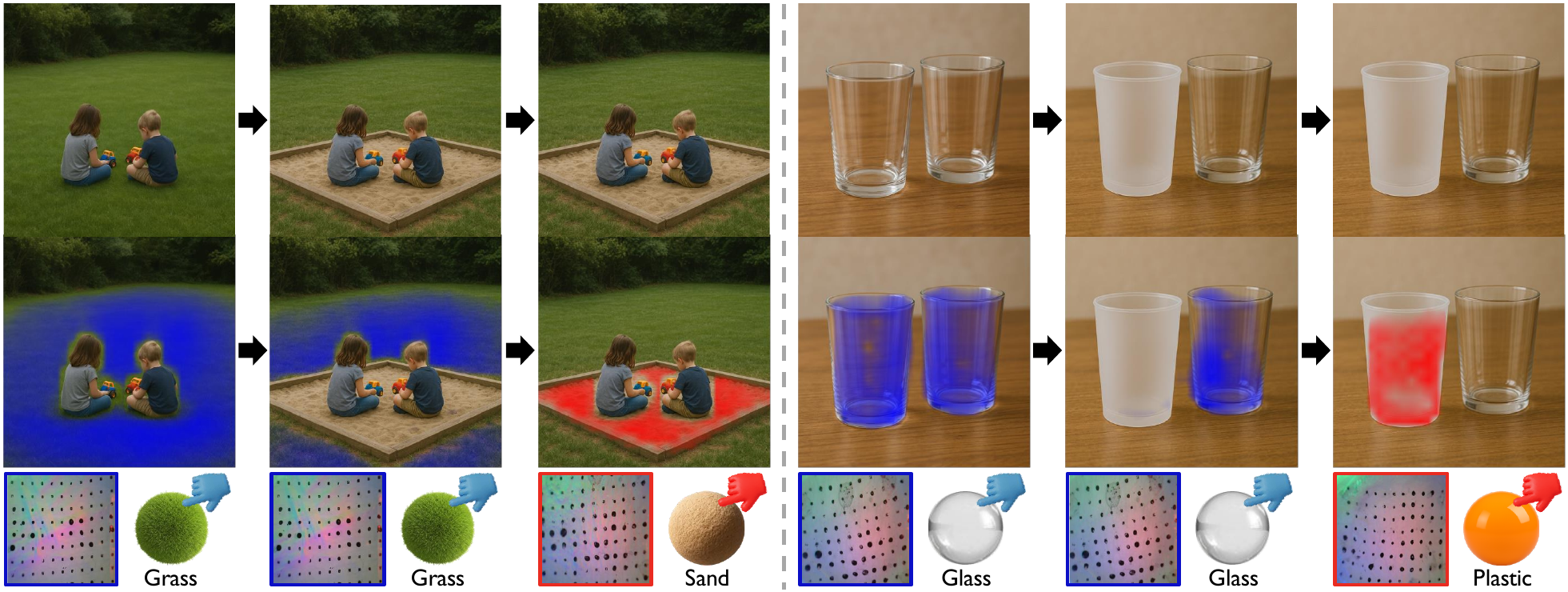}
    \caption{\textbf{Qualitative Results on Material Replacement.}}

    \label{fig:material_change_suppl}
\end{figure*}

\newpara{Material Replacement Results.} ~\Fref{fig:material_change_suppl} addresses the challenge of shape bias through material replacement scenarios. We compare original images with manipulated versions where specific regions are replaced with different materials while preserving the original object shape (\eg, a glass cup replaced by a plastic cup). Despite the identical geometric structure, our model accurately localizes the target regions according to their distinct tactile signals. This effectively validates that our method learns to localize based on material semantics (tactile properties) rather than relying on object shape.

\begin{figure*}[t!]
    \centering
    \includegraphics[width=0.95\linewidth]{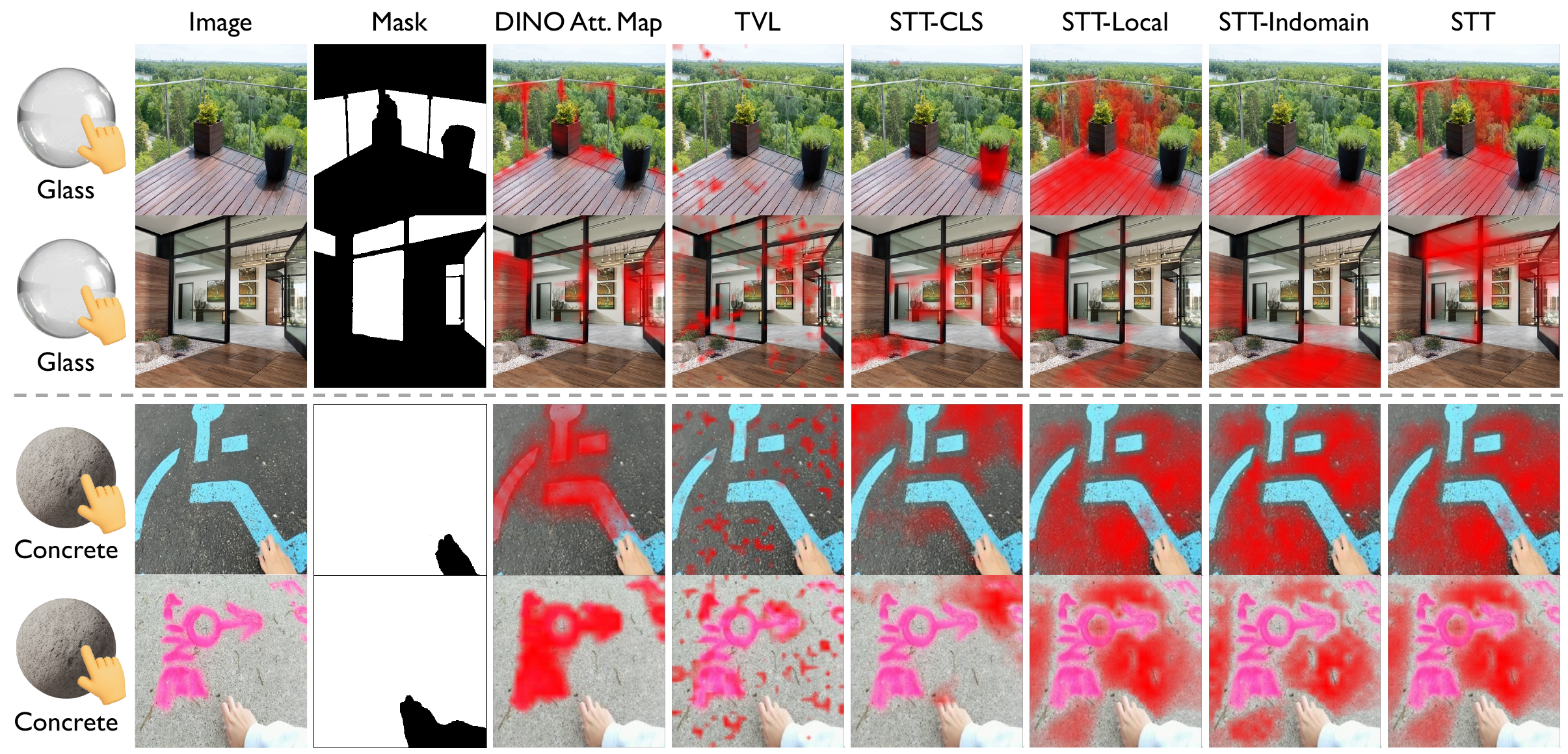}
    \caption{\textbf{Failure Cases.}}

    \label{fig:failure_cases}
\end{figure*}

\newpara{Failure Cases.} We analyze two representative failure cases in~\Fref{fig:failure_cases} to discuss the current limitations of our method.

\noindent \textit{Transparent Objects.} Segmenting transparent objects is a challenging problem due to their optical property of transmitting background objects and textures~\cite{xie2020segmenting}. Our model shares this limitation. As shown in our previous qualitative results, the model successfully segments glass objects or regions (relatively smaller in size) that exhibited internal refraction or specular highlights, making them easier to detect. However, as seen in the top two rows of~\Fref{fig:failure_cases}, the model struggles with larger highly transparent regions where the background texture is clearly visible, leading to inaccurate segmentation results. 

\noindent \textit{Painted Surface.} We observe another failure case when the target material is visually overlaid with high-contrast painted patterns, such as logos or symbols on concrete or asphalt surfaces. Although the underlying texture is largely preserved and implies a comparable tactile sensation, the model often suppresses these painted regions. This occurs because the visual appearance of the paint introduces strong color and edge cues that dominate the material’s true surface characteristics, leading the model to treat painted areas as a separate visual category. As a result, the model localizes only the unpainted concrete regions, despite the tactile equivalence across the entire surface.

\newpage



\stopcontents[supplementary]

\end{document}